\documentclass[10pt,twocolumn,letterpaper]{article}

\usepackage{iccv}
\usepackage{times}
\usepackage{epsfig}
\usepackage{graphicx}
\usepackage{amsmath}
\usepackage{amssymb}
\usepackage{mathtools}
\usepackage[ruled,vlined]{algorithm2e}
\usepackage{multirow}
\usepackage{subcaption}
\usepackage{setspace}
\usepackage{booktabs}
\usepackage{changepage}
\usepackage{wrapfig}
\usepackage{enumitem}
\usepackage{blindtext}
\usepackage{xcolor,colortbl}
\usepackage{footnote}
\definecolor{LightCyan}{rgb}{0.88,1,1}
\usepackage{appendix}

\usepackage{caption}

\usepackage{pifont}

\usepackage[pagebackref=true,breaklinks=true,letterpaper=true,colorlinks,bookmarks=false]{hyperref}

\iccvfinalcopy 

\def\iccvPaperID{6081} 

\ificcvfinal\fi

\title{SPG: Unsupervised Domain Adaptation for 3D Object Detection via \\ Semantic Point Generation}

\author{\hspace*{-29pt}Qiangeng Xu$^{1}$ $^\dagger$ \qquad Yin Zhou$^{2}$ \qquad Weiyue Wang$^{2}$ \qquad Charles R. Qi$^{2}$  \qquad Dragomir Anguelov$^{2}$ \\
    \hspace{-15mm}$^1$University of Southern California \hspace{30mm} $^2$Waymo, LLC\\
    {\tt\small \hspace{0mm}\{qiangenx\}@usc.edu}\hspace{5mm}{\tt\small \{yinzhou,weiyuewang,rqi,dragomir\}@waymo.com}\qquad
}
\begin{document}

\twocolumn[{%
\renewcommand\twocolumn[1][]{#1}%
\maketitle

\ificcvfinal\thispagestyle{empty}\fi

\begin{center}
    \vspace{-15pt}
    \centering
    \includegraphics[width=1\textwidth]{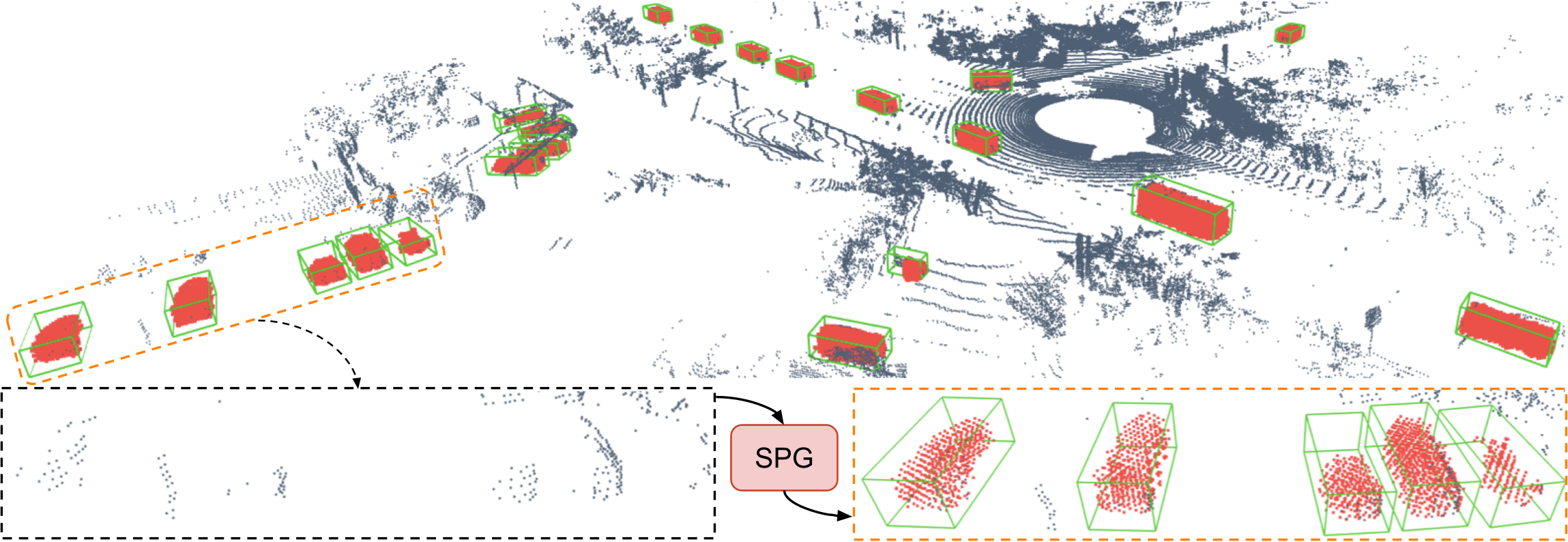}
    \captionof{figure}{Our Semantic Point Generation (SPG) recovers the foreground regions by generating semantic points (red). Combined with the original cloud, these semantic points can be directly used by modern LiDAR-based detectors and help improve the detection results (green boxes).}
\end{center}%
}]

\let\thefootnote\relax\footnotetext{\leftline{$^\dagger$Work done during internship at Waymo LLC. }}

\begin{abstract}

    In autonomous driving, a LiDAR-based object detector should perform reliably at different geographic locations and under various weather conditions. While recent 3D detection research focuses on improving performance within a single domain, our study reveals that the performance of modern detectors can drop drastically cross-domain. In this paper, we investigate unsupervised domain adaptation (UDA) for LiDAR-based 3D object detection. On the Waymo Domain Adaptation~\cite{sun2019scalability} dataset, we identify the deteriorating point cloud quality as the root cause of the performance drop. To address this issue, we present Semantic Point Generation (SPG), a general approach to enhance the reliability of LiDAR detectors against domain shifts.
    Specifically, SPG generates semantic points at the predicted foreground regions and faithfully recovers missing parts of the foreground objects, which are caused by phenomena such as occlusions, low reflectance or weather interference. By merging the semantic points with the original points, we obtain an augmented point cloud, which can be directly consumed by modern LiDAR-based detectors. 
    To validate the wide applicability of SPG, we experiment with two representative detectors, PointPillars~\cite{lang2019pointpillars} and PV-RCNN~\cite{shi2020pv}. 
    On the UDA task, SPG significantly improves both detectors across all object categories of interest and at all difficulty levels. 
    SPG can also benefit object detection in the original domain. On the Waymo Open Dataset~\cite{sun2019scalability} and KITTI~\cite{geiger2013vision}, SPG improves 3D detection results of these two methods 
    across all categories. Combined with PV-RCNN~\cite{shi2020pv}, SPG achieves state-of-the-art 3D detection results on KITTI.
    
\end{abstract}
\vspace{-15pt}
\section{Introduction}
A robust autonomous driving system requires its LiDAR-based detector to reliably handle different environmental conditions, \textit{e.g.}, geographic locations and weather conditions. 
While 3D detection has received increasing interest in recent years, most existing works~\cite{zhou2018voxelnet,chen2017multi,chen2019fast,chen2020dsgn,du2018general,konigshof2019realtime,lang2019pointpillars,li2019gs3d,li2019stereo,liang2019multi,liang2018deep,meyer2019lasernet,pon2020object,qi2018frustum,shi2020pv,shi2019pointrcnn,shi2019points,shi2020point,xu2020zoomnet,yan2018second,yang2018pixor,yang20203dssd,yang2019std,xu2020grid,zhou2020end} have focused on the performance in a single domain, where training and test data are captured in similar conditions. It is still an open question how to generalize a 3D detector to different domains, where the environment varies significantly.
In this paper, we address the domain gap caused by the deteriorating point cloud quality and aim to improve  3D object detection in the setting of unsupervised domain adaptation (UDA). We use the Waymo Domain Adaptation dataset \cite{sun2019scalability} to analyze the domain gap and introduce semantic point generation (SPG), a general approach to \textit{enhance the reliability of LiDAR detectors against domain shift}. SPG is able to improve detection quality in both the target domain and the source domain and can be naturally combined with modern LiDAR-based detectors.


\subsection{Understanding the Domain Gap}
\label{sec:dom_analysis}
	\textit{Waymo Open Dataset} (OD) is mainly collected in California and Arizona, and \textit{Waymo Kirkland Dataset} (Kirk) \cite{sun2019scalability} is collected in Kirkland. We consider OD as the source domain and Kirk as the target domain. To understand the possible domain gap, we take a PointPillars \cite{lang2019pointpillars} model trained on the OD training set and compare its 3D vehicle detection performance on OD validation set and those on Kirk validation set. We observe a drastic performance drop of $21.8$ points in 3D average precision (AP) (see Table \ref{tb:db_stats}).
	\begin{table}
        \begin{adjustwidth}{0pt}{0pt}
        \setlength\tabcolsep{3pt}
        \centering
		\begin{tabular}{c|c|c|c|c}
        \hline
        Dataset  & \begin{tabular}[c]{@{}c@{}}Rainy \\ frames\end{tabular} & \begin{tabular}[c]{@{}c@{}}Avg. number \\ of missing\\ points per frame\end{tabular} & \begin{tabular}[c]{@{}c@{}} Avg. number \\ of points \\ per vehicle\end{tabular} & \begin{tabular}[c]{@{}c@{}}3D L1 \\ AP\end{tabular} \\ \hline
        OD Val   & 0.5 \%                                                  & 23.0K                                                                   & 306.2                                                                 & 56.54                                                \\ \hline
        Kirk Dry & 0.0 \%                                                  & 25.1K                                                                   & 303.6                                                        & 55.98                                                \\ \hline
        Kirk Val & 100.0\%                                                 & 42.8K                                                                   & 222.3                                                                 & 34.74                                                \\ \hline
        \end{tabular}
	    \end{adjustwidth}
        \captionsetup{aboveskip=5pt}
        \captionsetup{belowskip=-5pt}
		\caption{The statistics of OD and Kirk. Each frame contains at most 163.8K points. Kirk Dry is formed by frames with dry weather in Kirk training set.}
		\label{tb:db_stats}
	\end{table}
    \begin{figure} 
        \begin{adjustwidth}{0pt}{-10pt}
            \begin{subfigure}{0.49\linewidth}
                \flushleft
                \includegraphics[width=0.99\linewidth]{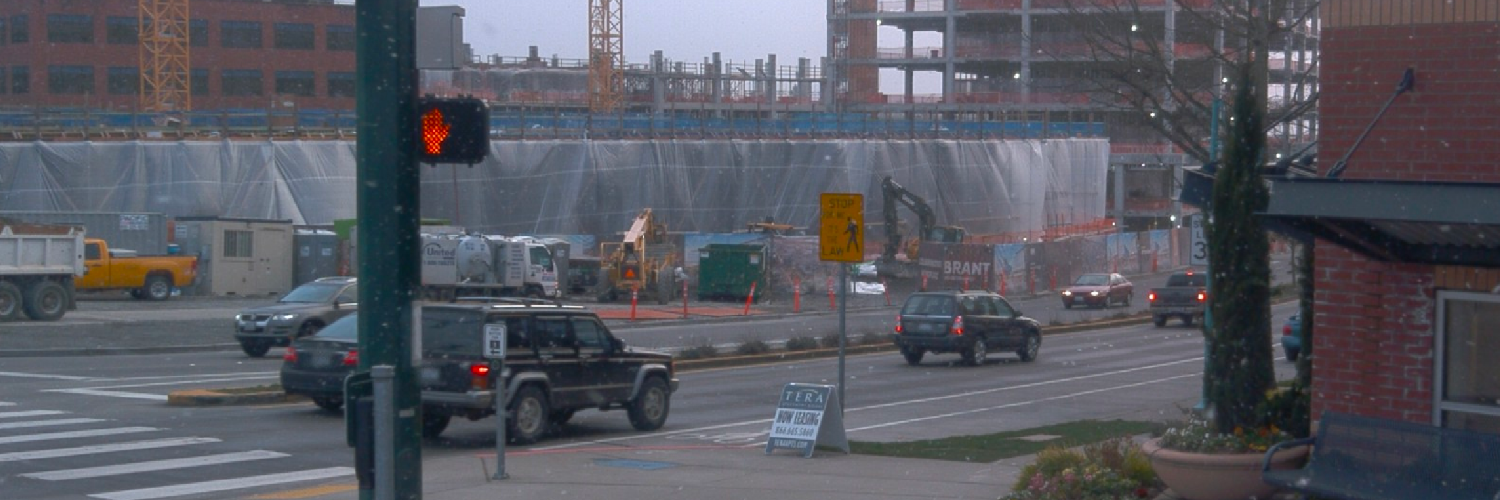}            \captionsetup{aboveskip = 1pt}
                \captionsetup{belowskip = 1pt}
                \caption{OD RGB Image}
            \end{subfigure}
            \begin{subfigure}{0.49\linewidth}
                \flushright
                \includegraphics[width=0.99\linewidth]{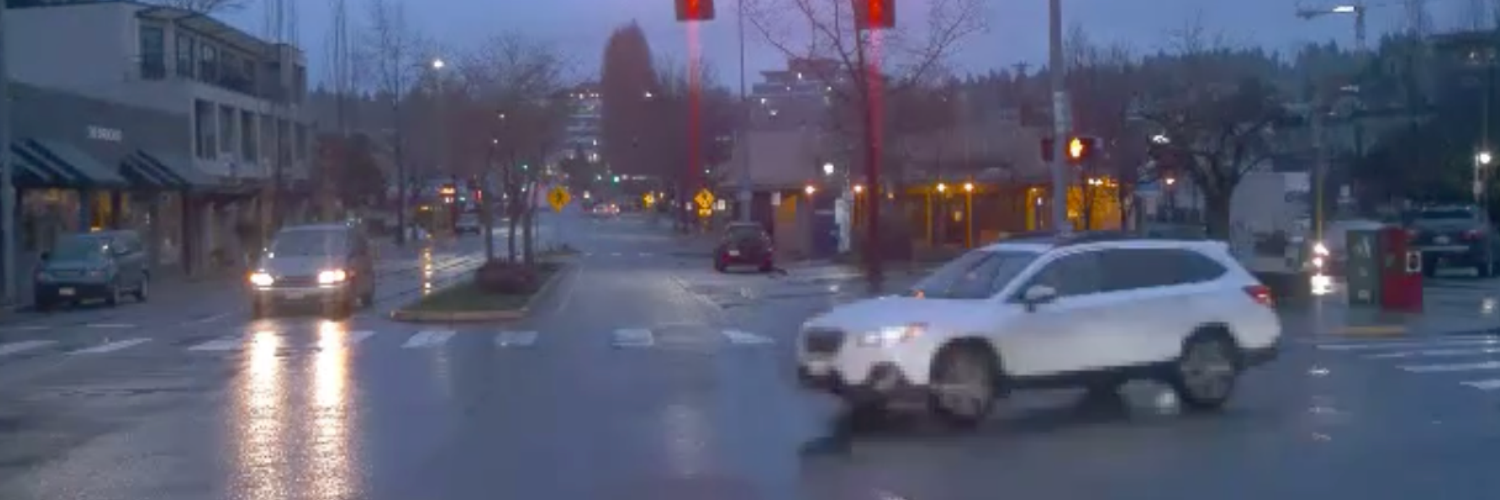}
                \captionsetup{aboveskip = 1pt}
                \captionsetup{belowskip = 1pt}
                \caption{Kirk RGB Image}
            \end{subfigure} 
            \begin{subfigure}{0.49\linewidth}
                \flushleft
                \includegraphics[width=0.99\linewidth]{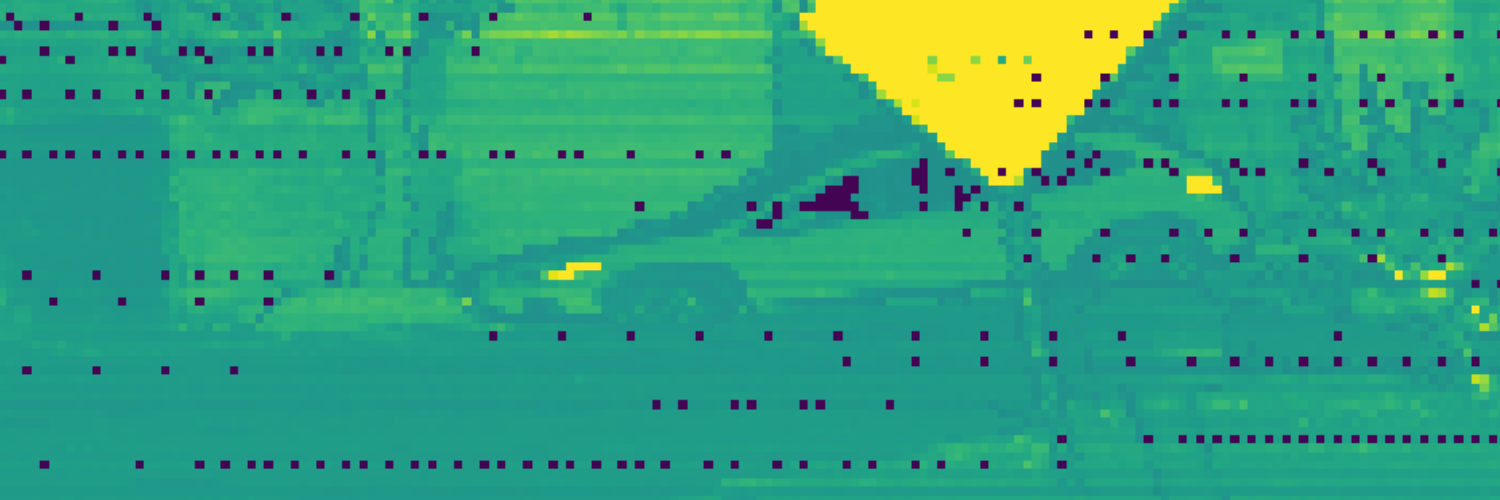}
                \captionsetup{aboveskip = 1pt}
                \captionsetup{belowskip = 1pt}
                \caption{OD Range Image}
            \end{subfigure}
            \begin{subfigure}{0.50\linewidth}
                \flushright
                \includegraphics[width=0.97\linewidth]{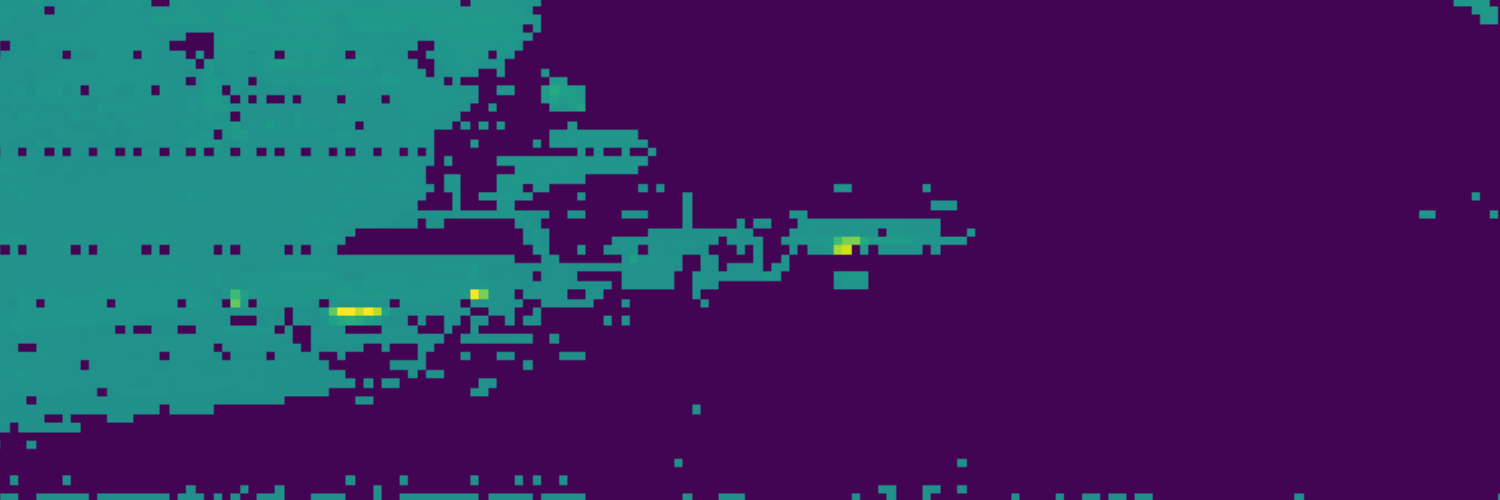}
                \captionsetup{aboveskip = 1pt}
                \captionsetup{belowskip = 1pt}
                \caption{Kirk Range Image}
            \end{subfigure}
        \captionsetup{aboveskip=5pt}
        \captionsetup{belowskip=-20pt}
        \caption{Examples of RGB and range image (intensity channel) in OD validation set and Kirk validation set. The dark regions in the range images indicate missed LiDAR returns. The regions of ``missing points" are irregular in shape.}
        \label{fig:range_missing}
        \end{adjustwidth}
    \end{figure}
    
    We first confirm that there is no significant difference in object size between two domains. Then by investigating the meta data in the datasets, we find that only $0.5\%$ of LiDAR frames in OD are collected under rainy weather, but almost all frames in Kirk share the rainy weather attribute. To rule out other factors, we extract all dry weather frames in Kirk training set and form a ``Kirk Dry" dataset. Because the the rain drop changes the surface property of objects, there are twice amount of missing LiDAR points per frame in Kirk validation set than in OD or Kirk Dry (see Table \ref{tb:db_stats}). As a result, vehicles in Kirk receive around $27\%$ fewer LiDAR point observations than those in OD (see statistics and more details in the supplemental). In Figure \ref{fig:range_missing}, we visualize two range images from OD and Kirk, respectively. We can observe that in the rainy weather, a significant number of points are missing and the distribution of missing points is more irregular compared to the dry weather. 
    
    To conclude, the major domain gap between OD and Kirk is the deteriorating point cloud quality, which is caused by the rainy weather condition. In the target domain, we name this phenomenon as the ``\textbf{missing point}" problem.
    
    \subsection{Previous Methods to Address the Domain Gap}
    Multiple studies propose to align the features across domains. Most of them focus on 2D tasks \cite{morerio2017minimal,ganin2015unsupervised,tzeng2017adversarial,dong2019semantic} or object-level 3D tasks \cite{zhou2018unsupervised,qin2019pointdan}. Applying feature alignment \cite{chen2018domain,he2019multi,luo2020unsupervised} requires a redesign of the model or loss of a detector. Our goal is to seek a general solution to benefit recently reported LiDAR-based detectors\cite{lang2019pointpillars,shi2020pv,zhou2018voxelnet,shi2019pointrcnn,he2020sassd}. 
    
    Another direction is to apply transformations to the data from one domain to match the data from another domain. A naive approach is to randomly down-sample the point cloud but this not only fails to satisfactorily simulate the pattern of missing points (Figure \ref{fig:range_missing}d) but also hurts the performance on the source domain. Another approach is to up-sample the point cloud~\cite{yu2018pu,yifan2019patch,li2019pu} in the target domain, which can increase point density around observed regions. However, those methods have a limited capability in recovering the 3D shape of very partially observed objects. Moreover, up-sampling the entire point cloud will lead to a significantly higher latency. A third approach is to leverage style transfer techniques: \cite{zhu2017unpaired,park2020contrastive,choi2019self,he2019multi,shan2019pixel,hsu2020progressive,saleh2019domain} render point clouds as 2D pseudo images and enforce the renderings from different domains to be resemblant in style. However, these methods introduce an information bottleneck during rasterization ~\cite{zhou2018voxelnet} and they are not applicable to modern point-based 3D detectors~\cite{shi2020pv}.
    
    
    
   
   \subsection{SPG for Closing the Domain Gap}
   The ``missing point'' problem deteriorates the point cloud quality and reduces the number of point observations, thus undermining the detection performance. To address this issue, we propose Semantic Point Generation (SPG). Our approach aims to learn the semantic information of the point cloud and performs foreground region prediction to identify voxels that are inside foreground objects. Based on the predicted foreground voxels, SPG generates points to recover the foreground regions. Since these points are discriminatively generated at foreground objects, we denote them by \textbf{semantic points}. These semantic points are merged with the original points into an augmented point cloud, which is then fed to a 3D detector. 
   
   The contributions of this paper are two-fold: \\
   1. We present an in-depth analysis of unsupervised domain adaptation (UDA) for LiDAR 3D detectors across different geographic locations and weather conditions. Our study reveals that the rainy weather can severely deteriorate the quality of LiDAR point clouds and lead to drastic performance drop for modern detectors. \\
   2. We propose semantic point generation (SPG). To our best knowledge, it is the first learning-based model that targets UDA for point cloud 3D detection. Specifically, SPG has the following merits:
   \begin{itemize}[noitemsep, topsep=2pt, leftmargin=8pt]
   \item SPG can generate semantic points that faithfully recover the foreground regions suffering from the ``missing point'' problem. SPG can significantly improve performance over poor-quality point clouds in the target domain while also benefiting source domain, for representative 3D detectors, including PointPillars \cite{lang2019pointpillars} and PV-RCNN \cite{shi2020pv}.  
   \item SPG also improves the performance for the general 3D object detection task. We verify its effectiveness on KITTI \cite{geiger2013vision} for the aforementioned 3D detectors.
   \item SPG is a general approach and can be easily combined with modern off-the-shelf LiDAR-based detectors. 
   \item Our approach is light-weight and efficient. Introducing less than $6\%$ additional points, SPG only adds a marginal complexity to a 3D detector.  
   \end{itemize}
\section{Related Work}
    \subsection{Unsupervised Domain Adaptation}
Unsupervised domain adaptation (UDA) aims to generalize a model to a novel (target) domain by using label information only from the source domain.
The two domains are generally related, but there exists a distribution shift (domain gap). Most methods focus on learning aligned feature representations across domains. To reach this goal, \cite{borgwardt2006integrating} proposes Maximum Mean Discrepancy (MMD) while \cite{pan2010domain} proposes Transfer Component Analysis (TCA). \cite{long2013transfer} designs a Joint Distribution Adaptation to close the distribution shift while \cite{long2015learning,long2016unsupervised} utilize a shared Hilbert space. Without using explicit distance measures, deep learning models  \cite{ganin2015unsupervised,tzeng2017adversarial,dong2019semantic,qin2019generatively,saito2018maximum} use adversarial training to get indistinguishable features between domains.

\vspace{-10pt}
\paragraph{Unsupervised Domain Adaptation for 2D Detection}
The object detection task is sensitive to local geometric features. \cite{chen2018domain,he2019multi} hierarchically align the features between domains. Most of these works focus on UDA for 2D detection. With the current advances of unpaired style transfer methods \cite{park2020contrastive,zhu2017unpaired}, studies such as \cite{shan2019pixel,hsu2020progressive} translate the image from source domain to target domain or vice versa. 

\vspace{-10pt}
\paragraph{Unsupervised Domain Adaptation for 3D Tasks} Most of the UDA methods focus on 2D tasks, only a few studies explore the UDA in 3D. \cite{zhou2018unsupervised,qin2019pointdan} align the global and local features for object-level tasks. To reduce the sparsity, \cite{wu2019squeezesegv2} projects the point cloud to 2D view, while \cite{saleh2019domain} projects the point cloud to birds-eye view (BEV). \cite{du2020associate} creates a car model set and adapts their features to the detection object features. However, this study targets general car 3D detection on a single point cloud domain. \cite{wang2020train} is the first published study targeting UDA for 3D LiDAR detection. They identify the vehicle size as the domain gap between KITTI\cite{geiger2013vision} and other datasets. So they resize the vehicles in the data. In contrast, we identify the point cloud quality as the major domain gap between Waymo's two datasets\cite{sun2019scalability}. We use a learning-based approach to close the domain gap. 

\subsection{Point Cloud Transformation}
One way to improve point cloud quality is to suitably transform the point cloud. Studies of point cloud up-sampling \cite{yu2018pu,yifan2019patch,li2019pu} can transfer 
a low density point cloud to a high density one.
However, they need high density point cloud ground truth during training. These networks can densify the point cloud in the observed regions. But in our case, we also need to recover regions with no point observation, caused by ``missing points''.

Point cloud completion networks \cite{yuan2018pcn,chen2019unpaired,yang2018foldingnet,xie2020grnet} aim to complete the point cloud. Specialized in object-level completion, these models assume a single object has been manually located and the input only consists of the points on this object. Therefore, these models do not fit our purpose of object detection. Point cloud style transfer models \cite{cao2020psnet,cao2019neural} can transfer the color theme and the object-level geometric style for the point cloud. However, these models do not focus on preserving local details with high-fidelity. Therefore, their transformation cannot directly help 3D detection.
\section{Semantic Point Generation}
        \begin{figure*} 
    \vspace{-10pt}
        \begin{adjustwidth}{0pt}{0pt}
            \centering
            \includegraphics[width=0.96\linewidth]{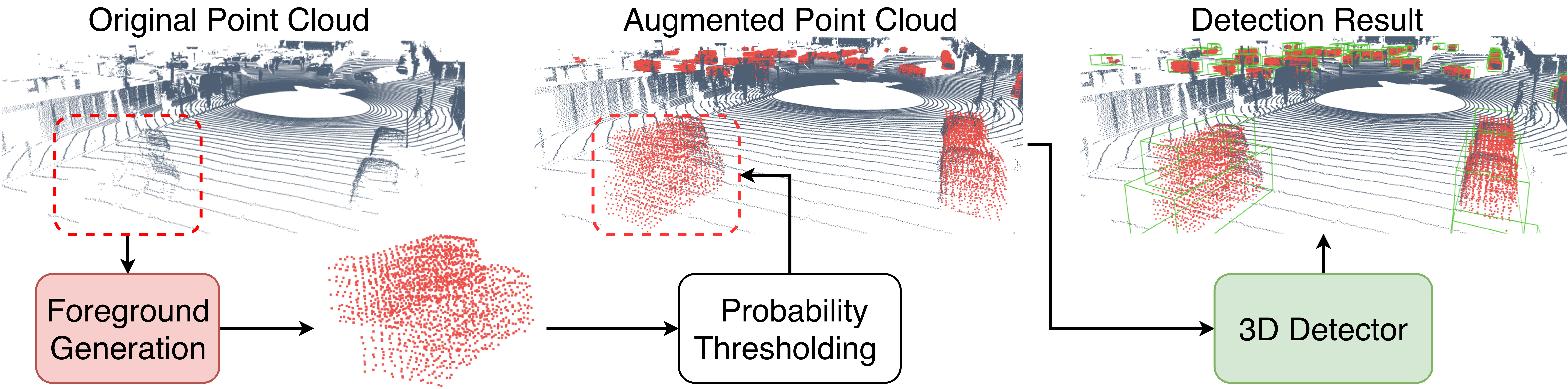}
            \captionsetup{aboveskip=5pt}
            \captionsetup{belowskip=-10pt}
            \caption{Illustration of SPG-aided 3D detection pipeline. SPG voxelizes the entire point cloud and generates prediction for each voxel (both occupied and empty) within the generation areas. After applying a probability thresholding, we take the top voxels with highest foreground probability and add a semantic point (red) at the predicted location in each of these voxels. These points are merged with the original point cloud and fed into the selected 3D point cloud detector.}
            \label{fig:Method_Overview}
        \end{adjustwidth}
    \end{figure*}
    
    In the input point cloud $PC_{raw} = \{p_1, p_2, ..., p_N \} \in \mathbb{R} ^{3+F}$, each point has three channels of xyz and $F$ properties (\textit{e.g.}, intensity, elongation). Figure \ref{fig:Method_Overview} illustrates the SPG-aided 3D detection pipeline. SPG takes raw point cloud $PC_{raw}$ as input and generates a set of semantic points
    in the predicted foreground regions. Then, these semantic points are combined with the original point cloud into an augmented point cloud $PC_{aug}$, which is fed into a point cloud detector to obtain object detection results.
    
    As shown in Figure \ref{fig:Network_Architecture}, SPG voxelizes $PC_{raw}$ into an evenly spaced 3D voxel grid, and learns the point cloud semantics for these voxels. For each voxel, the network predicts the probability confidence $\tilde{P}^{f}$ of it being a foreground voxel (contained in a foreground object bounding box). In each foreground voxel, the network generates a \textbf{semantic point} $\tilde{sp}$ with point features $\tilde{\psi}=[\tilde{\chi}, \tilde{f}]$. $\tilde{\chi} \in \mathbb{R}^{3}$ is the xyz coordinate of $\tilde{sp}$ and $\tilde{f} \in \mathbb{R}^F$ is the point properties.
    
    To faithfully recover the foreground regions of the observed objects, we define a \textbf{generation area}. Only voxels occupied or neighbored by the observed points are considered within the generation area. We also filter out semantic points with $\tilde{P}^{f}$ less than $P_{thresh}$, then take $K$ semantic points $\{\tilde{sp}_1, \tilde{sp}_2, ..., \tilde{sp}_K \}$ with the highest $\tilde{P}^{f}$ and merge them with the original point cloud $PC_{raw}$ to get $PC_{aug}$. In practice, we use $P_{thresh}=0.5$.
    
    To enable SPG to be directly used by modern LiDAR-based detectors, we encode the augmented point cloud $PC_{aug}$ as $\{\hat{p}_1, \hat{p}_2, ..., \hat{p}_N, \tilde{sp}_1, \tilde{sp}_2, ..., \tilde{sp}_K \} \in \mathbb{R} ^{3+F+1}$. Here we add another property channel to each point, indicating the confidence in foreground prediction: $\tilde{P}^{f}$ is used for the semantic points, and 1.0 for the original raw points. 
    
    \subsection{Training Targets}
        To train SPG, we need to create two supervisions: 1) $y^{f}$, the class label if a voxel (either occupied or empty) is a foreground voxel, which supervises $\tilde{P}^{f}$; 2) $\psi \in \mathbb{R} ^{3+F}$, the regression target for semantic point features $\tilde{\psi}$.
        
\begin{figure*}
        \vspace{-15pt}
            \begin{adjustwidth}{0pt}{0pt}
                \centering
                \includegraphics[width=0.95\linewidth]{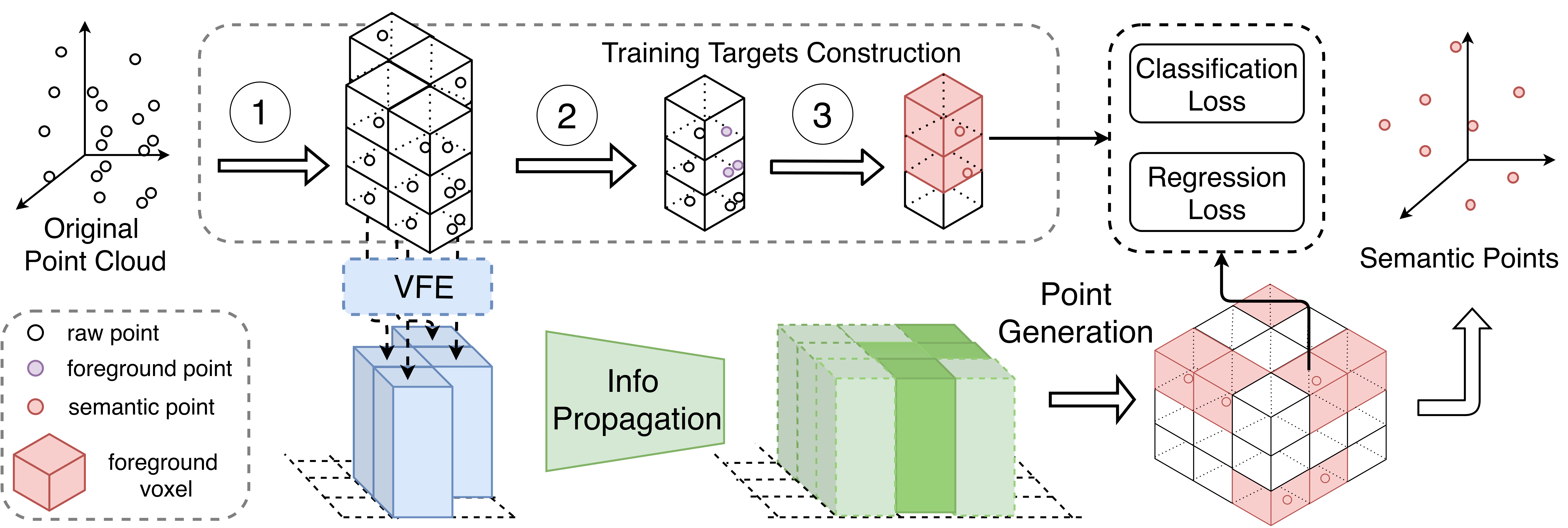}
                \captionsetup{aboveskip=5pt}
                \captionsetup{belowskip=-10pt}
                \caption{Training targets construction and SPG model architecture. Three steps to create the semantic point training targets: 1.Voxelization; 2. Foreground points searching 3. Label assignment and ground-truth point feature calculation. 
                SPG includes: the Voxel Feature Encoding module (VFE), the Information Propagation module, and the Point Generation module.}
                \label{fig:Network_Architecture}
            \end{adjustwidth}
        \end{figure*}

        As visualized in Figure \ref{fig:Network_Architecture}, we mark a point as a foreground point if it is inside an object bounding box. Voxels contained in a foreground bounding box are marked as foreground voxels $V^f$. For voxel $v_i$, we assign $y^{f}_i = 1$ if $v_i \in V^f$ and $y^{f}_i = 0$ otherwise. 
        If $v_i$ is an occupied foreground voxel, we set $\psi_i = [\bar{\chi}_{i}, \bar{f}_i]$ as the regression target, where $\bar{\chi}_{i} \in \mathbb{R} ^3$ is the centroid (xyz) of all foreground points in $v_i$ while $\bar{f}_i \in \mathbb{R}^F$ is the mean of their point properties (e.g. intensity, elongation).
        
    \subsection{Model Structure}
        The lower part of Figure \ref{fig:Network_Architecture} illustrates the network architecture. SPG uses a light-weight encoder-decoder network \cite{zhou2018voxelnet, lang2019pointpillars}, which is composed of three modules: \\
        1) The Voxel Feature Encoding module~\cite{zhou2018voxelnet} aggregates points inside each voxel by using several MLPs. Similar to \cite{lang2019pointpillars, shi2020pv}, these voxel features are later stacked into pillars and projected onto a birds-eye view feature space; \\
        2) The Information Propagation module applies 2D convolutions on the pillar features. As shown in Figure \ref{fig:Network_Architecture}, the semantic information in the occupied pillars (dark green) is populated into the neighboring empty pillars (light green), which enables SPG to recover the foreground regions in the empty space.  \\ 
        3. The Point Generation module maps the pillar features to the corresponding voxels. 
        For each voxel $v_i$ in the generation area, the module creates a semantic point $\tilde{sp}_i$ with encoding $[\tilde{\chi}_i, \tilde{f}_i, \tilde{P}^{f}_i]$, in which $\tilde{\chi_i}$ is the point location, $\tilde{f}_i$ is the point properties, and $\tilde{P}^{f}_i$ is the foreground probability.
        
	\subsection{Foreground Region Recovery}
	    The above pipeline supervises SPG to generate semantic points in the occupied voxels. However, it is also crucial to recover the empty voxels caused by the ``missing points'' problem. To generate semantic points in the empty areas, SPG employs two strategies:
    	\begin{itemize}[noitemsep, topsep=2pt, leftmargin=8pt]
           \item ``Hide and Predict", which produces the ``missing points'' on the source domain during training and guides SPG to recover the foreground object shape in the empty space.
           \item ``Semantic Area Expansion", which leverages the foreground/background voxel labels derived from the bounding boxes and encourages SPG to recover more unobserved foreground regions in each bounding box.
        \end{itemize}
        \vspace{-10pt}
        \subsubsection{Hide and Predict}
        SPG voxelizes $PC_{raw} \in \mathbb{R} ^{3+F}$ into a voxel set $V = \{v_1, v_2, ..., v_M \}$. Before passing $V$ to the network, we randomly select $\gamma\%$ of the occupied voxels $V_{hide} \subset V$ and hide all their points. 
        During training, SPG is required to predict the foreground/background label $y^f$ for all voxels in $V$, even though it only observes points in $|V - V_{hide}|$. The predicted point features $\tilde{\psi}$ in $V_{hide}^f$ should match the corresponding ground-truth $\psi$ calculated by these hidden points.
        
        This strategy brings two benefits: 1. Hiding points region by region mimics the missing point pattern in the target domain; 2. The strategy naturally creates the training targets for semantic points in the empty space. Section \ref{sec:abs} shows the effectiveness of this strategy. Here we set $\gamma = 25$.
        \vspace{-20pt}
        \subsubsection{Semantic Area Expansion}
        \label{sec:sae}
            \begin{figure} 
                \setlength{\abovedisplayskip}{0pt}%
                \setlength{\abovedisplayshortskip}{\abovedisplayskip}%
                \begin{adjustwidth}{0pt}{0pt}
                    \centering
                    \includegraphics[width=1.0\linewidth]{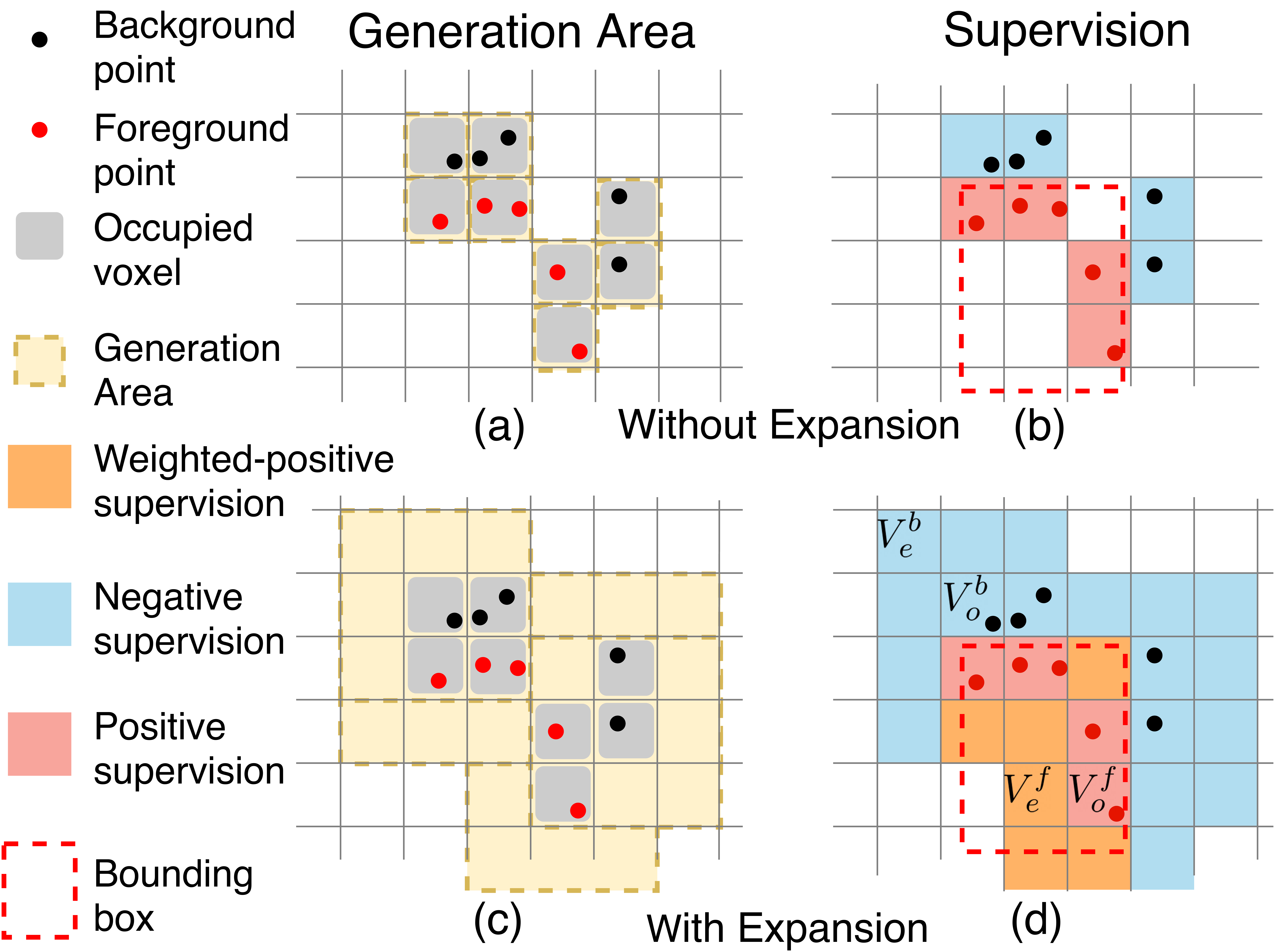}
                    \captionsetup{aboveskip=5pt}
                    \captionsetup{belowskip=-25pt}
                    \caption{Visualization of ``Semantic Area Expansion''. (a) and (c) show the occupied voxels and the generation area, respectively. (b) and (d) show the supervision strategies.}
                    \label{fig:xpsion}
                \end{adjustwidth}
            \end{figure}
           In section \ref{sec:dom_analysis}, we find the poor point cloud quality leads to insufficient points on each object and substantially degrades the detection performance. To remedy this problem, we allow SPG to expand the generation area to the empty space. Figure \ref{fig:xpsion} a and c show the examples of the generation area with and without the expansion, respectively. 
            
            Without the expansion, we can use the ground-truth knowledge of foreground points to supervise SPG only on the occupied voxels (Figure \ref{fig:xpsion} b). However, with the expansion, there is no foreground point inside these empty voxels. Therefore, as shown in Figure \ref{fig:xpsion} d, we design a supervision scheme as follows: \\
            1. For both occupied and empty background voxels $V_o^{b}$ and $V_e^{b}$, we impose negative supervision and set label $y^{f}= 0$. \\
            2. For the occupied foreground voxels $V_o^{f}$, we set $y^{f}= 1$. \\
            3. For the empty voxels inside a bounding box $V_e^{f}$ , we set their foreground label $y^{f}= 1$ and assign a weighting factor $\alpha$, where $\alpha<1$. \\
            4. We only impose point features supervision $\psi$ at occupied foreground voxels $V_o^{f}$. 
            
\begin{figure}
                    \setlength{\abovedisplayskip}{0pt}%
                    \setlength{\belowdisplayskip}{0pt}%
                    \begin{adjustwidth}{0pt}{0pt}
                        \begin{subfigure}{0.49\linewidth}
                            \flushleft
                            \includegraphics[width=0.93\linewidth]{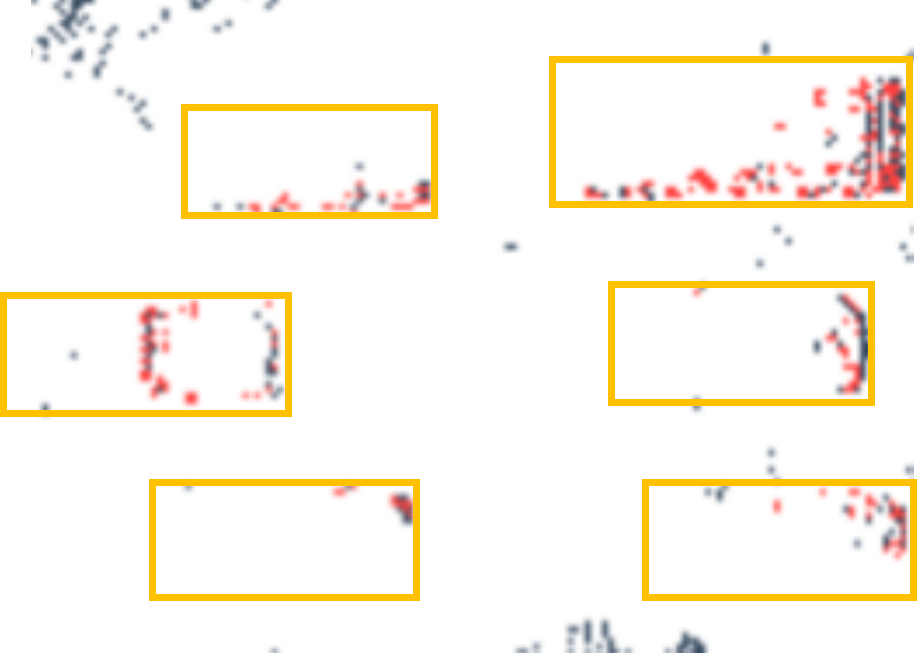}
                            \caption{Without expansion}
                        \end{subfigure}
                        \begin{subfigure}{0.49\linewidth}
                            \flushright
                            \includegraphics[width=0.95\linewidth]{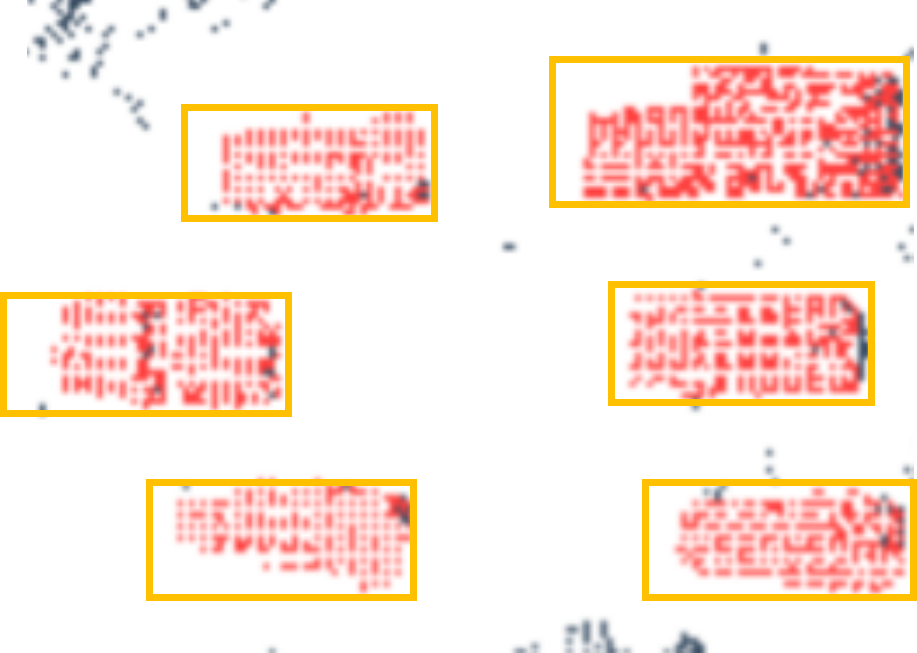}
                            \caption{With expansion}
                        \end{subfigure}
                    \captionsetup{aboveskip=5pt}   
                    \captionsetup{belowskip=-15pt}
                    \caption{Comparisons between generated semantic points (red) with and without ``Semantic Area Expansion''.}
                    \label{fig:xpsion_result}
                \end{adjustwidth}
            \end{figure}

            To investigate the effectiveness of the expansion, we train a model on the OD training set and evaluate it on the Kirk validation set. The expansion results in $\textbf{510\%}$ more semantic points on foreground objects, which mitigates the ``missing points" problem caused by environmental interference and occlusions. 
            Figure \ref{fig:xpsion_result} shows the generation results with and without the expansion. The supervision scheme encourages SPG to learn the extended shape of vehicle parts and enables SPG to fill in more foreground space with semantic points. We also conduct ablation studies (Section \ref{sec:abs}) to show the effectiveness of the proposed strategy.
    \subsection{Objectives}
        We use two loss functions, \textit{i.e.,} foreground area classification loss $L_{cls}$ and feature regression loss $L_{reg}$. 
    
        To supervise $\tilde{P}^f$ with label $y^f$, we use Focal loss \cite{lin2017focal} to mitigate the background-foreground class imbalance. $L_{cls}$ can be decomposed as focal losses on four categories of voxels: the occupied voxels $V_o$, the empty background voxels $V_e^b$, the empty foreground voxels $V_e^f$ and the hidden voxels $V_{hide}$. The labeling strategy for these categories is described in Section \ref{sec:sae}. 
        \begin{adjustwidth}{0pt}{0pt}
            \setlength{\abovedisplayskip}{-5pt}%
            \setlength{\abovedisplayshortskip}{\abovedisplayskip}%
            \setlength{\belowdisplayskip}{5pt}%
            \begin{align}
            L_{cls} &= \frac{1}{|V_o \cup V_e^b|}\sum\nolimits_{V_o \cup V_e^b}{L_{focal}} \nonumber \\ 
            &+ \frac{\alpha}{|V_e^f|} \sum\nolimits_{V_e^f}{L_{focal}} + \frac{\beta}{|V_{hide}|} \sum\nolimits_{V_{hide}}{L_{focal}}
            \end{align}
        \end{adjustwidth} 
        We use Smooth-L1 loss \cite{he2019multi} for point feature $\tilde{\psi}$ regression, and supervise on the semantic points in occupied foreground voxels $V_o^f$ and the hidden foreground voxels $V_{hide}^f$.
         \begin{adjustwidth}{0pt}{0pt}
            \setlength{\abovedisplayskip}{-5pt}%
            \setlength{\abovedisplayshortskip}{\abovedisplayskip}%
            \setlength{\belowdisplayskip}{5pt}%
            \begin{align}
            L_{reg} &= \frac{1}{|V_o^f|}\sum\nolimits_{V_o^f}{L_{smooth-L1}(\tilde{\psi}, \psi)} \nonumber \\
                    &+ \frac{\beta}{|V_{hide}^f|} \sum\nolimits_{V_{hide}^f}{L_{smooth-L1}(\tilde{\psi}, \psi)}
            \end{align}
        \end{adjustwidth} 
        Please note that we are only interested in the $L_{cls}$ and $L_{reg}$ on voxels inside the generation area. We find $\alpha = 0.5$ and $\beta = 2.0$ achieves the best result.
\vspace{-5pt}
\section{Experiments}
        In this section, we first evaluate the effectiveness of SPG as a general UDA approach for 3D detection, based on the Waymo Domain Adaptation Dataset \cite{sun2019scalability}. In addition, we show that SPG can also improve results for top-performing 3D detectors on the source domain\cite{sun2019scalability,geiger2013vision}.
    To demonstrate the wide applicability of SPG, we choose two representative detectors: 1) PointPillars \cite{lang2019pointpillars}, popular among industrial-grade autonomous driving systems; 2) PV-RCNN \cite{shi2020pv}, a high performance LiDAR-based 3D detector ~\cite{geiger2013vision,sun2019scalability}. We perform two groups of model comparisons under the setting of unsupervised domain adaptation (UDA) and general 3D object detection: group 1, PointPillars vs. SPG + PointPillars; group 2, PV-RCNN vs. SPG + PV-RCNN. SPG can also be combined with range image-based detectors \cite{meyer2019lasernet,zhou2020end,REF:Range_AlexBewley2020} by applying ray casting to the generated points. However, we leave this as future work. 
    \vspace{-13pt}
    \paragraph{Datasets} The Waymo Domain Adaptation dataset 1.0~\cite{sun2019scalability} consists of two sub datasets, the \textit{Waymo Open Dataset} (OD) and the \textit{Waymo Kirkland Dataset} (Kirk). OD provides 798 training segments of 158,361 LiDAR frames and 202 validation segments of 40,077 frames. Captured across California and Arizona, $99.40\%$ of its frames have dry weather. Kirk is a smaller dataset including 80 training segments of 15,797 frames and 20 validation segments of 3,933 frames. Captured in Kirkland, $97.99\%$ its LiDAR frames have rainy weather. 
    To examine a detector's reliability when entering a new environment, we conduct UDA experiments without using the data in Kirk during training.
    
    \textit{KITTI}~\cite{geiger2013vision} 
    contains 7481 training samples and 7518 testing samples. Following~\cite{REF:Multiview3D_2017}, we divide the training data into a \textit{train} split and a \textit{val} split containing 3721 and 3769 LiDAR frames, respectively.
	\vspace{-13pt}
    \paragraph{Implementation and Training Details}
    We use a single lightweight network architecture on all experiments. As shown in Figure \ref{fig:Network_Architecture}, our Voxel Feature Encoding\cite{zhou2018voxelnet} module includes a single layer point-wise MLP and a voxel-wise max-pooling \cite{qi2017pointnet,zhou2018voxelnet}. The  Information Propagation module includes two levels of CNN layers. The first level includes three CNN layers with stride 1. The second level includes one CNN layer with stride 2 and four subsequent CNN layers with stride 1, then up-sampled back to the original resolution. Each layer has an output dimension of 128. From the BEV feature map, the Point Generation module uses one FC layer to produce $\tilde{P}^f$ and another FC layer to generate the features $\tilde{\psi}$ for the voxels in each pillar. SPG and each detector are trained separately. 
    
    We implement PointPillars following~\cite{lang2019pointpillars} and use the PV-RCNN code provided by \cite{shi2020pv} (the training settings on OD 1.0 are obtained via direct communication with the author). 
    On the Waymo Domain Adaptation Dataset~\cite{sun2019scalability}, we set the voxel dimensions to (0.32m, 0.32m, 0.4m) for PointPillars and (0.2m, 0.2m, 0.3m) for PV-RCNN. On KITTI, we set the voxel dimensions to (0.16m, 0.16m, 0.2m) and (0.2m, 0.2m, 0.3m) for PointPillars and PV-RCNN, respectively. By default, the generation area includes voxels within 6 steps of any occupied voxel. After probability thresholding, we preserve up to $8000$ semantic points for the Waymo Domain Adaptation Dataset and $6000$ for KITTI.
    
        \begin{table*}[hbt]
            \vspace{-10pt}
            \begin{adjustwidth}{0pt}{0pt}
            \setlength\tabcolsep{6.5pt}
            \begin{tabular}{cccccccccc}
            \multicolumn{1}{l}{}                           & \multicolumn{1}{l}{}                                       & \multicolumn{4}{c}{Target Domain - Kirk}                                                                                                                                        & \multicolumn{4}{c}{Source Domain - OD}                                                                                                                     \\ \hline
            \multicolumn{1}{c|}{}                          & \multicolumn{1}{c|}{}                                      & \multicolumn{2}{c|}{Vehicle}                                                           & \multicolumn{2}{c|}{Pedestrian}                                                        & \multicolumn{2}{c|}{Vehicle}                                                           & \multicolumn{2}{c}{Pedestrian}                                   \\
            \multicolumn{1}{c|}{Difficulty}                & \multicolumn{1}{c|}{Method}                                & 3D AP                           & \multicolumn{1}{c|}{BEV AP}                          & 3D AP                           & \multicolumn{1}{c|}{BEV AP}                          & 3D AP                           & \multicolumn{1}{c|}{BEV AP}                          & 3D  AP                          & BEV AP                          \\ \hline
            \multicolumn{1}{c|}{\multirow{3}{*}{LEVEL\_1}} & \multicolumn{1}{c|}{PointPillars}                          & 34.65                           & \multicolumn{1}{c|}{51.88}                           & 20.65                           & \multicolumn{1}{c|}{22.33}                           & 57.27                           & \multicolumn{1}{c|}{72.26}                           & 55.20                           & 63.82                           \\
            \multicolumn{1}{c|}{}                          & \multicolumn{1}{c|}{SPG + PointPillars}                    & \textbf{41.56} & \multicolumn{1}{c|}{\textbf{60.44}} & \textbf{23.72} & \multicolumn{1}{c|}{\textbf{24.83}} & \textbf{62.44} & \multicolumn{1}{c|}{\textbf{77.63}} & \textbf{56.06} & \textbf{64.66} \\\rowcolor{LightCyan}
            \multicolumn{1}{c|}{}                          & \multicolumn{1}{c|}{\textit{Improvement}} & \textit{+6.91} & \multicolumn{1}{c|}{\textit{+8.56}} & \textit{+3.07} & \multicolumn{1}{c|}{\textit{+2.50}} & \textit{+5.17} & \multicolumn{1}{c|}{\textit{+5.37}} & \textit{+0.86} & \textit{+0.84} \\ \hline
            \multicolumn{1}{c|}{\multirow{3}{*}{LEVEL\_2}} & \multicolumn{1}{c|}{PointPillars}                          & 31.67                           & \multicolumn{1}{c|}{47.93}                           & 17.66                           & \multicolumn{1}{c|}{18.40}                           & 52.96                           & \multicolumn{1}{c|}{69.09}                           & 51.33                           & 60.13                           \\
            \multicolumn{1}{c|}{}                          & \multicolumn{1}{c|}{SPG + PointPillars}                    & \textbf{38.15} & \multicolumn{1}{c|}{\textbf{56.94}} & \textbf{19.57} & \multicolumn{1}{c|}{\textbf{20.67}} & \textbf{58.54} & \multicolumn{1}{c|}{\textbf{74.90}} & \textbf{52.33} & \textbf{60.93} \\\rowcolor{LightCyan}
            \multicolumn{1}{c|}{}                          & \multicolumn{1}{c|}{\textit{Improvement}} & \textit{+6.48} & \multicolumn{1}{c|}{\textit{+9.01}} & \textit{+1.91} & \multicolumn{1}{c|}{\textit{+2.27}} & \textit{+5.58} & \multicolumn{1}{c|}{\textit{+5.81}} & \textit{+1.00} & \textit{+0.80} \\ \hline\hline
            \multicolumn{1}{c|}{\multirow{3}{*}{LEVEL\_1}} & \multicolumn{1}{c|}{PV-RCNN}                               & 55.16                           & \multicolumn{1}{c|}{70.38}                           & 24.47                           & \multicolumn{1}{c|}{25.39}                           & 74.01                           & \multicolumn{1}{c|}{85.13}                           & 65.34                           & 70.35                           \\
            \multicolumn{1}{c|}{}                          & \multicolumn{1}{c|}{SPG + PV-RCNN}                         & \textbf{58.31} & \multicolumn{1}{c|}{\textbf{72.56}} & \textbf{30.82} & \multicolumn{1}{c|}{\textbf{31.92}} & \textbf{75.27} & \multicolumn{1}{c|}{\textbf{87.38}} & \textbf{66.93} & \textbf{70.37} \\\rowcolor{LightCyan}
            \multicolumn{1}{c|}{}                          & \multicolumn{1}{c|}{\textit{Improvement}} & \textit{+3.15} & \multicolumn{1}{c|}{\textit{+2.18}} & \textit{+6.35} & \multicolumn{1}{c|}{\textit{+6.53}} & \textit{+1.26} & \multicolumn{1}{c|}{\textit{+2.25}} & \textit{+1.59} & \textit{+0.02} \\ \hline
            \multicolumn{1}{c|}{\multirow{3}{*}{LEVEL\_2}} & \multicolumn{1}{c|}{PV-RCNN}                               & 45.81                           & \multicolumn{1}{c|}{60.13}                           & 17.16                           & \multicolumn{1}{c|}{17.88}                           & 64.69                           & \multicolumn{1}{c|}{76.84}                           & 56.03                           & 60.81                           \\
            \multicolumn{1}{c|}{}                          & \multicolumn{1}{c|}{SPG + PV-RCNN}                         & \textbf{48.70} & \multicolumn{1}{c|}{\textbf{62.03}} & \textbf{22.05} & \multicolumn{1}{c|}{\textbf{22.65}} & \textbf{65.98} & \multicolumn{1}{c|}{\textbf{78.05}} & \textbf{57.68} & \textbf{60.88} \\\rowcolor{LightCyan}
            \multicolumn{1}{c|}{}                          & \multicolumn{1}{c|}{\textit{Improvement}} & \textit{+2.89} & \multicolumn{1}{c|}{\textit{+1.90}} & \textit{+4.89} & \multicolumn{1}{c|}{\textit{+4.77}} & \textit{+1.29} & \multicolumn{1}{c|}{\textit{+1.21}} & \textit{+1.65} & \textit{+0.07} \\ \cline{1-10} 
            \end{tabular}
            \captionsetup{aboveskip = 5pt}
            \captionsetup{belowskip = -15pt}
    		\caption{Results on the Waymo Open Dataset 1.0 and the Kirkland Dataset. Results for PointPillars are based on our own implementation following \cite{lang2019pointpillars}. We use the PV-RCNN source code and obtain training settings for the Waymo Open Dataset~\cite{sun2019scalability} via direct communication with the author.}
    		\label{tb:uda}
    		\end{adjustwidth}
		\end{table*}
    \subsection{Evaluation on the Waymo Open Dataset}
    \label{sec:Exp_UDA}
        We perform two groups of model comparisons by training them on the OD training set and evaluating them on both the OD validation set and the Kirk validation set. 
        \vspace{-10pt}
        \paragraph{Evaluation Metrics} The Kirk 1.0 validation set only provides the evaluation labels for the vehicle and the pedestrian classes. We use the official evaluation tool released by \cite{sun2019scalability}. The IoU thresholds for vehicles and pedestrians are 0.7 and 0.5. In Table \ref{tb:uda} we report both 3D and BEV AP on two difficulty levels. More results with distance breakdown are shown in the supplemental material.
        \vspace{-10pt}
        \paragraph{Target Domain}
        On Kirk, we observe that SPG brings remarkable improvements over both detectors across all object types. Averaged over two difficulty levels, SPG improves PointPillars on Kirk vehicle 3D AP by $6.7\%$ and BEV AP by $8.8\%$. For PV-RCNN, SPG improves Kirk pedestrian 3D AP by $5.6\%$ and BEV AP by $5.7\%$. 
        \vspace{-10pt}
        \paragraph{Source Domain}
        Unlike most UDA methods \cite{chen2018domain,hsu2020progressive,shan2019pixel} that only optimize the performance on the target domain, SPG also consistently improves the results on the source domain. Averaged across both difficulty levels, SPG improves OD vehicle 3D AP for PointPillars by $5.4\%$ and improves OD pedestrian 3D AP for PV-RCNN by $1.6\%$.
        \vspace{-10pt}
        \paragraph{Comparison with Alternative Strategies} We compare SPG with alternative strategies that also target the deteriorating point cloud quality. We employ PointPillars as the baseline and choose LEVEL\_1 vehicle 3D AP as the main metric on the Kirk validation set, during UDA. Three strategies are implemented: 1. RndDrop, where we randomly drop $17\%$ of the points in the source domain during training. This dropout ratio is chosen 
        for the number of points in the source and target domain to match (see Table \ref{tb:db_stats}). 2. K-frames, where we use $K$ consecutive historical frames in both the source domain and the target domain. The points in the first $K-1$ are transformed into the last frame according to the ground-truth ego-motion, so that the last frame has $K$ times the number of points. 3. Adversarial Domain Adaptation (ADA), where we follow \cite{ganin2015unsupervised} and add a domain classification loss on the pillar features of PointPillars.
        
    	As shown in Table~\ref{tb:strtgy}, although ``RndDrop" enforces the quantity of missing points in the source domain to match with that in the target domain, the pattern of missing points still differs from the reality (see Figure \ref{fig:range_missing}), which limits the improvement to only $0.80\%$ in 3D AP. To remedy the ``missing points'' problem, ``3-frames'' contains real points from 3 frames and ``5-frames'' contains points from 5 frames. With around 800K points per scene, ``5-frames'' significantly improves the single-frame baseline. However, aggregating multiple frames inevitably 
    	increases the memory usage and the processing time. ADA improves 3D AP to $36.34$ on the target domain, but we observe an AP drop of $1.52$ in the source domain. Remarkably, SPG can outperform ``5-frames'', by adding only 8000 semantic points, which is less than $6\%$ of the points in a single frame.
    	
    	\begin{table}[!hbt]
            \setlength\tabcolsep{1.3pt}
            \begin{tabular}{c|c|c|c|c|c|c}
                \hline
                Method   & Baseline & RndDrop & 3-frames & 5-frames & ADA & SPG            \\ \hline
                3D AP & 34.65    & 35.45    & 38.00   & 38.51  & 36.34 & \textbf{41.56} \\ \hline
                \end{tabular}
            \captionsetup{aboveskip = 5pt}
            \captionsetup{belowskip = -5pt}
    		\caption{Comparisons of different strategies targeting at the deteriorating point cloud quality. The models are trained on OD and evaluated on Kirk. The metric is LEVEL\_1 Vehicle 3D AP. We use PointPillars\cite{lang2019pointpillars} as the baseline.}
    		\label{tb:strtgy}
		\end{table}
    	
    \subsection{Evaluation on the KITTI Dataset}
		In this section, we show besides the usefulness in UDA (Sec. \ref{sec:Exp_UDA}) the proposed SPG can also boost performance in another popular 3D detection benchmark (i.e. KITTI~\cite{geiger2013vision}). 
		We follow the training and evaluation protocols in~\cite{lang2019pointpillars,shi2020pv}.
		\begin{table} 
            \vspace{-10pt}
            \begin{adjustwidth}{0pt}{0pt}
            \setlength\tabcolsep{2.8pt}
                \begin{tabular}{c|c|cccc}
                \hline
                            &           & \multicolumn{4}{c}{Car - 3D AP}                                  \\
                Method      & Reference & Easy           & Mod.           & Hard           & Avg.           \\ \hline
                SA-SSD\cite{he2020sassd}      & \small CVPR 2020 & 88.75          & 79.79          & 74.16          & 80.90          \\
                3D-CVF\cite{yoo20203d}      & \small ECCV 2020 & 89.20          & 80.05          & 73.11          & 80.79          \\
                CIA-SSD\cite{zheng2020ciassd}     & \small AAAI 2021 & 89.59          & 80.28          & 72.87          & 80.91          \\
                Asso-3Ddet\cite{du2020associate}  & \small CVPR 2020 & 85.99          & 77.40          & 70.53          & 77.97         \\   
                Voxel R-CNN\cite{deng2020voxel} & \small AAAI 2021 & \textbf{90.90} & 81.62          & 77.06          & 83.19          \\ \hline
                PV-RCNN\cite{shi2020pv}     & \small CVPR 2020 & 90.25          & 81.43          & 76.82          & 82.83          \\
                \textbf{SPG}+PV-RCNN  & -         & 90.50          & \textbf{82.13} & \textbf{78.90} & \textbf{83.84} \\ \hline
                \end{tabular}
            \captionsetup{aboveskip = 5pt}
            \captionsetup{belowskip = -15pt}
    		\caption{Car detection Results on the KITTI test set. See the full list of comparisons in the supplemental.}
    		\label{tb:bench}
            \end{adjustwidth}
        \end{table}
        \vspace{-15pt}
        \paragraph{KITTI Test Set} As shown in Table \ref{tb:bench}, SPG significantly improves PV-RCNN on Car 3D detection. As of Mar. 3rd, 2021, our method ranks the \textbf{1st} on KITTI car 3D detection among all published methods (4th among all submitted approaches). 
        Moreover, SPG demonstrates strong robustness in detecting hard objects (truncation up to 50\%). Specifically, SPG surpasses all submitted methods on the hard category by a big margin and achieves the \textbf{highest} overall 3D AP of $83.84\%$ (averaged over Easy, Mod. and Hard).
        
        \begin{table*}[hbt]
            \vspace{-5pt}
            \begin{adjustwidth}{0pt}{0pt}
            \setlength\tabcolsep{5pt}
           \begin{tabular}{c|ccc|ccc|ccc|ccc}
            \hline
           & \multicolumn{3}{c|}{Car - 3D AP}                                                                    & \multicolumn{3}{c|}{Car - BEV AP}                                                                   & \multicolumn{3}{c|}{Pedestrian - 3D AP}                                                             & \multicolumn{3}{c}{Pedestrian - BEV AP}                                                            \\
            Method                                                                                 & Easy                            & Mod.                            & Hard                            & Easy                            & Mod.                            & Hard                            & Easy                            & Mod.                            & Hard                            & Easy                            & Mod.                            & Hard                            \\ \hline
            PointPillars                                                                           & 87.75                           & 78.39                           & 75.18                           & 92.03                           & 88.05                           & 86.66                           & 57.30                           & 51.41                           & 46.87                           & 61.59                           & 56.01                           & 52.04                           \\
            SPG + PointPillars                                                                     & \textbf{89.77} & \textbf{81.36} & \textbf{78.85} & \textbf{94.38} & \textbf{89.92} & \textbf{87.97} & \textbf{59.65} & \textbf{53.55} & \textbf{49.24} & \textbf{65.38} & \textbf{59.48} & \textbf{55.32} \\
            \rowcolor{LightCyan}            \textit{Improvement} & \textit{+2.02}                            & \textit{+2.97}                            & \textit{+3.67}                            & \textit{+2.35}                            & \textit{+1.87}                            & \textit{+1.31}                            & \textit{+2.35}                            & \textit{+2.14}                            & \textit{+2.47}                            & \textit{+3.79}                            & \textit{+3.47}                            & \textit{+3.28}                            \\ \hline
            PV-RCNN                                                                                & 92.10                           & 84.36                           & 82.48                           & 93.02                           & 90.33                           & 88.53                           & 64.26                           & 56.67                           & 51.91                           & 67.97                           & 60.52                           & 55.80                           \\
            SPG + PV-RCNN                                                                          & \textbf{92.53} & \textbf{85.31} & \textbf{82.82} & \textbf{94.99} & \textbf{91.11} & \textbf{88.86} & \textbf{69.66} & \textbf{61.80} & \textbf{56.39} & \textbf{71.79} & \textbf{64.50} & \textbf{59.51} \\
            \rowcolor{LightCyan}            \textit{Improvement} & \textit{+0.43}                            & \textit{+0.95}                            & \textit{+0.34}                            & \textit{+1.97}                            & \textit{+0.78}                            & \textit{+0.33}                            & \textit{+5.40}                            & \textit{+5.13}                            & \textit{+4.48}                            & \textit{+3.82}                            & \textit{+3.98}                            & \textit{+3.71}                            \\ \hline
            \end{tabular}
            \captionsetup{aboveskip = 5pt}
            \captionsetup{belowskip = -15pt}
    		\caption{Comparisons on the KITTI validation set. Average Precision (AP) is computed over 40 recall positions. The baseline results\cite{shi2020pv,openpcdet2020} are obtained based on publically released models. See more results (including Cyclist) in the supplemental.
    		}
    		\label{tb:kitti}
    		\end{adjustwidth}
		\end{table*}
        \vspace{-10pt}
        \paragraph{KITTI Validation Set} We summarize the results in Table \ref{tb:kitti}. We train each group of models using the recommended settings of baseline detectors \cite{lang2019pointpillars,shi2020pv}.  
        
        SPG remarkably improves both PointPillars and PV-RCNN on all object types and difficulty levels. Specifically, for PointPillars, SPG improves the 3D AP of car detection by $2.02\%$, $2.97\%$, $3.67\%$ on easy, moderate, and hard levels, respectively. 
        For PV-RCNN, SPG improves the 3D AP of pedestrian detection by $5.40\%$, $5.13\%$, $4.48\%$ on easy, moderate and hard levels, respectively.
    \vspace{0pt}
    \subsection{Model Efficiency} We evaluate the efficiency of SPG on the KITTI \textit{val} split (Table \ref{tb:eff}). SPG contains $0.39$ million parameters while adding less than $17$ milliseconds latency to the detectors. This indicates that SPG is highly efficient for industrial-grade deployment on a stringent computation budget. 
		\begin{table}[hbt]
            \begin{adjustwidth}{0pt}{0pt}
            \setlength\tabcolsep{4pt}
            \begin{tabular}{c|cc|cc|c}
                \hline
                Detectors            & \multicolumn{2}{c|}{PointPillars} & \multicolumn{2}{c|}{PV-RCNN} & -     \\ \hline
                With SPG             & No              & Yes             & No            & Yes          & Yes   \\ \hline
                Latency (ms)         & 23.56           & 36.67           & 139.96        & 156.85       & 16.82 \\ \hline
                Parameters& 4.83M            & 5.22M            & 13.12M         & 13.51M        & 0.39M  \\ \hline
            \end{tabular}
            \captionsetup{aboveskip = 5pt}
            \captionsetup{belowskip = -5pt}
    		\caption{Latency and model parameters. ``M'' stands for million. The last column shows the results of standalone SPG. The evaluation is based on a 1080Ti GPU with a batch size of 1. The latency is averaged over the KITTI \textit{val} split.}
    		\label{tb:eff}
    		\end{adjustwidth}
		\end{table}
		
    \vspace{-5pt}
    \subsection{Ablation Studies}
    \label{sec:abs}
    	\begin{table}[hbt]
    	    \vspace{-5pt}
            \setlength\tabcolsep{1.9pt}
            \begin{tabular}{l|c|c|c|c|c}
                \hline
                          &                  & Hide \& & Foreground  & 3D             &                         \\
                Model     & Expansion        & Predict & Confidence & AP             & \textit{Improve}        \\ \hline
                Baseline  & $-$                & $-$       & $-$           & 34.65          & \textit{$-$}              \\
                SPG       & $-$                & $-$    &$\checkmark$             & 35.89          & \textit{+1.24}          \\
                SPG       & $-$                & $25\%$    &$\checkmark$             & 38.09          & \textit{+3.44}          \\
                SPG & $\checkmark$($\alpha$=0.0) & $25\%$    & $\checkmark$         & 38.96          & \textit{+4.31}          \\
                SPG & $\checkmark$($\alpha$=1.0) & $25\%$    & $\checkmark$         & 38.42          & \textit{+3.77}          \\
                SPG & $\checkmark$($\alpha$=0.5) & $-$       & $\checkmark$         & 39.22          & \textit{+4.57}          \\
                SPG       & $\checkmark$($\alpha$=0.5) & $25\%$    & $-$           & 37.96          & \textit{+3.31}          \\
                SPG(ours) & $\checkmark$($\alpha$=0.5) & $25\%$    & $\checkmark$         & \textbf{41.56} & \textit{\textbf{+6.91}} \\ \hline
            \end{tabular}
            \captionsetup{aboveskip = 5pt}
            \captionsetup{belowskip = -10pt}
    		\caption{Ablation studies of SPG. The models are trained on OD and evaluated on Kirk. The metric is LEVEL\_1  Vehicle 3D AP. We use PointPillars\cite{lang2019pointpillars}  as our baseline.}
    		\label{tb:ablation}
		\end{table}
		We conduct ablation studies on ``Semantic Area Expansion'', ``Hide and Predict'' and whether to add foreground confidence ($\tilde{P}^f$) as a point property and show all of them can benefit detection quality (see Table \ref{tb:ablation}). We also change the weighting factor $\alpha$ on the empty foreground voxels $V_e^f$. A larger $\alpha$ encourages more point generation in the empty foreground space. However, in reality, an object typically does not occupy the entire space within a bounding box. Therefore, over-aggressively generating points does not help improve the performance (see $\alpha=1.0$). 
    	\vspace{-10pt}
    	\paragraph{Probability Thresholding} In Table~\ref{tb:tresh}, we show the effect of choosing different thresholds during probability thresholding. While a higher $P_{thresh}$ only keeps semantic points with high foreground probability, a lower $P_{thresh}$ admits more points, 
    	but may introduce points to the background. We find the threshold of $0.5$ achieves the best results.
		\begin{table} 
            \setlength\tabcolsep{7pt}
            \begin{tabular}{c|c|c|c|c|c}
                \hline\rule{0pt}{10pt}
                $P_{thresh}$ & $0.3$  & $0.4$  & $0.5$   & $0.6$   & $0.7$   \\ \hline\rule{0pt}{10pt}
                3D AP                 & 39.39 & 40.09 & \textbf{41.56} & 41.18 & 40.89 \\ \hline
            \end{tabular}
            \captionsetup{aboveskip = 5pt}
            \captionsetup{belowskip = -15pt}
    		\caption{Ablation studies on the probability threshold $P_{thresh}$ (only keep the semantic point if $\tilde{P}^{f} >$ $P_{thresh}$). Our best SPG model uses $P_{thresh}=0.5$. The metric is LEVEL\_1 Vehicle 3D AP on the Kirk validation set.}
    		\label{tb:tresh}
		\end{table}
\section{Conclusions}
    In this paper, we investigate unsupervised domain adaptation for LiDAR-based 3D detectors across different geographic locations and weather conditions. We observe that rainy weather can severely deteriorate the point cloud quality and cause drastic performance drop for modern 3D detectors, based on the Waymo Domain Adaptation dataset. The proposed SPG method addresses this issue as a novel unsupervised domain adaptation (UDA) task without using any training data from the new domain. This setting allows us to rigorously test 3D detectors against real-world challenges autonomous vehicles may experience due to diverse conditions (e.g., different levels of fog/rain/snow beyond what one may effectively train for) during the trip.

Utilizing two strategies ``Hide and Predict'' and ``Semantic Area Generation'', SPG generates semantic points to recover the shape of foreground objects with a negligible overhead (only adding $6\%$  extra points) and can be conveniently integrated with modern LiDAR-based detectors. We test SPG with two detectors: PointPillars and PV-RCNN. For unsupervised domain adaptation, SPG achieves significant performance gains on the challenging target domain. On Waymo Open dataset and KITTI, SPG also consistently benefits detection quality on the source domain.

    
\section{Acknowledgement}
We would like to thank Boqing Gong for the helpful discussions. We also thank Jingwei Ji for the careful proofreading.     
    
{\small
\bibliographystyle{ieee_fullname}
\bibliography{egbib}
}
\begin{appendices}
    \clearpage
    \pagebreak
    In this supplementary material, we provide detailed analysis about the statistics of the Waymo Domain Adaptation Dataset in Section \ref{sec:Detailed Analysis of the Domain Gap in the Waymo Domain Adaptation Dataset}; the robustness analysis of the foreground voxel classifier in Section \ref{sec:The Robustness of Foreground Voxel Classification}; the derivation of the dropout rate used in the RndDrop method in Section \ref{sec:Drop Rate of the RndDrop Method}; more results on the Waymo Domain Adaptation Dataset in Section \ref{sec:More Results on Waymo Domain Adaptation Dataset}; more results on KITTI in Section \ref{sec:More Results on KITTI}; and more visualization of the semantic point generation in Section \ref{sec:More Visualization of Semantic Point Generation}.

\section{Statistics of the Waymo Domain Adaptation Dataset}
    \label{sec:Detailed Analysis of the Domain Gap in the Waymo Domain Adaptation Dataset}
    \begin{figure}[!htb]
        \begin{adjustwidth}{0pt}{0pt}
            \centering
            \includegraphics[width=1.0\linewidth]{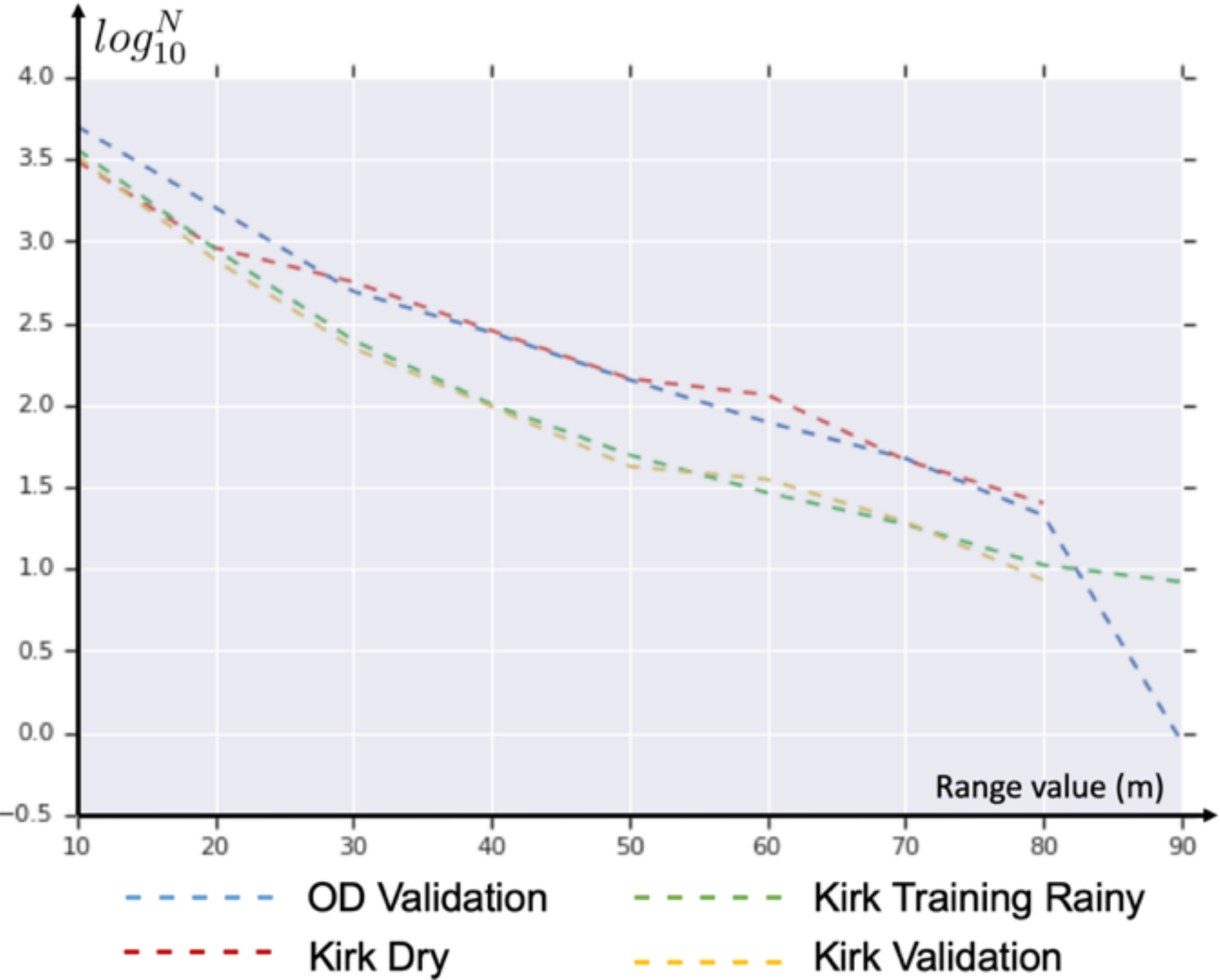}
            \caption{The average number of raw points per vehicle across different ranges.  On the x axis, the range value stands for the distance between the center of a bounding box and the LiDAR sensor. The y axis shows the value after applying $log_{10}$ on the number of points $N$. ``Kirk Dry'' is extracted from the Kirk Training set and contains frames captured in the dry weather.}
            \label{fig:num_pts_x_ranges}
        \end{adjustwidth}
    \end{figure}
    We collect the statistics about the average number of points in a vehicle bounding box across different ranges. The range value is calculated as the euclidean distance between the LiDAR sensor and the center of a bounding box. We investigate four sets of point clouds: 
    \begin{itemize}[noitemsep, topsep=2pt, leftmargin=8pt]
       \item The OD Validation set, in which $99.5\%$ of the frames are collected in the dry weather.
       \item The Kirk Dry set, which consists of all the frames with the dry weather condition from the Kirk training set. 
       \item The Kirk Training Rainy set, which consists of all the frames with the rainy weather condition from the Kirk training set.
       \item The Kirk Validation set, in which all the frames are collected in the rainy weather.
   \end{itemize}
    
    As shown in Figure \ref{fig:num_pts_x_ranges}, the point clouds with similar weather conditions share similar numbers of points per object, even though they are collected at different locations. Specifically, the vehicle objects of the two ``dry datasets'', \textit{i.e.,} the Kirk Dry set and the OD Validation set, have similar numbers of points across all ranges. The vehicle objects of the two ``rainy datasets'' \textit{i.e.,} the Kirk Training Rainy set and the Kirk Validation set, share similar statistics.
    
   In addition, the point clouds captured in \textbf{the dry weather} (the OD Validation set and the Kirk Dry set) have more points on each object than those collected in \textbf{the rainy weather} (the Kirk Training Rainy set and the Kirk Validation set). Please note that we have applied $log_{10}$ to the number of points for better visualization. The difference in the number of points is substantial between two weather conditions across all ranges. 
    
\section{The Robustness of the Foreground Voxel Classifier}
    \label{sec:The Robustness of Foreground Voxel Classification}
    In order to generalize detectors to different domains, it is crucial to correctly classify foreground voxels so that semantic points can be reliably generated. Table \ref{tb:fap_gap} lists the evaluation results of the foreground voxel classifier. 
    \begin{table}[!htb]
        \setlength\tabcolsep{3pt}
		\begin{tabular}{c|c|cccc}
            \hline
            Train       & Eval      & Accuracy & Precision & Recall  & AP     \\ \hline
            OD Train             & OD Val            & 99.3 \%  & 90.9 \%   & 92.9 \% & 86.7 \% \\ \hline
            OD Train            & Kirk Val           & 98.9 \%  & 88.4 \%   & 88.2 \% & 78.3 \% \\ \hline
        \end{tabular}
		\caption{Foreground voxel classification results of our SPG. The model is trained on the OD training set and then it is evaluated on the OD validation set and Kirk validation set, respectively. The accuracy, precision and recall are evaluated by setting $\tilde{P}^{f} > 0.5$.}
		\label{tb:fap_gap}
	\end{table}
	The results in Table \ref{tb:fap_gap} are averaged among all voxels in the foreground regions. Our SPG is trained on the OD training set. Then it is evaluated on the OD validation set and the Kirk validation set, respectively. The classification of a voxel is correct if its prediction score $\tilde{P}^{f} > 0.5$ when $y^f = 1.0$ or $\tilde{P}^{f} < 0.5$ when $y^f = 0.0$. The accuracy, precision and recall are all calculated under this setting. The AP is calculated using 40 recall thresholds. The results show that SPG achieves high performance in both domains. 
\section{Dropout Rate of the RndDrop Method}
    \label{sec:Drop Rate of the RndDrop Method}
    In the experiment section, we implement a baseline method RndDrop, where we randomly drop out $17\%$ of points for point clouds from the source domain during training. This dropout ratio is chosen to match the ratio of missing points in the target domain. We calculate $(\overline{N}_{src} - \overline{N}_{tgt}) / \overline{N}_{src} = 17\%$, where $\overline{N}_{src} = 121.2 K $ is the average number of points per scene in the source domain and $\overline{N}_{tgt} = 100.4 K $ is the average number of points per scene in the target domain.  

\section{More Results on the Waymo Domain Adaptation Dataset}
    \label{sec:More Results on Waymo Domain Adaptation Dataset}
    
    The evaluation tool \cite{sun2019scalability} provides the average precision for three distance-based breakdowns: 0 to 30 meters, 30 to 50 meters, and beyond 50 meters. The AP is calculated using 100 recall thresholds.
    
    We perform two groups of model comparisons in the setting of UDA: Group 1. PointPillars vs. SPG + PointPillars; Group 2. PV-RCNN vs. SPG + PV-RCNN. We train all models on the OD training set and evaluate them on both the OD validation set and the Kirk validation set. Table \ref{tb:uda_breakdown_v3d} and \ref{tb:uda_breakdown_vbev} show the comparisons on vehicle 3D AP and vehicle BEV AP, respectively. Table \ref{tb:uda_breakdown_p3d} and Table \ref{tb:uda_breakdown_pbev} show the comparisons in pedestrian 3D AP and pedestrian BEV AP, respectively. In most cases, SPG improves the detection performance across all ranges for both vehicles and pedestrians.
    
    \begin{table*}[!hbt]
      \begin{adjustwidth}{0pt}{0pt}
      \setlength\tabcolsep{7pt}
      \centering
       \begin{tabular}{c|c|cccc|cccc}
        \multicolumn{1}{l}{}                           & \multicolumn{1}{l}{}                                       & \multicolumn{4}{c}{Target Domain - Kirk}                                                                                                                                        & \multicolumn{4}{c}{Source Domain - OD}                                                                                                                     \\ 
        \hline
                                  &                      & \multicolumn{4}{c|}{Vehicle 3D AP (IoU = 0.7)}             & \multicolumn{4}{c}{Vehicle 3D AP (IoU = 0.7)}               \\
        Difficulty                & Method               & Overall        & 0-30m          & 30-50m         & 50-Inf         & Overall        & 0-30m          & 30-50m         & 50-Inf         \\ \hline
        \multirow{3}{*}{LEVEL\_1} & PointPillars          & 34.65          & 63.13          & 24.56          & 7.65           & 57.27          & 84.39          & 52.97          & 28.22          \\
                                  & SPG + PointPillars      & \textbf{41.56} & \textbf{68.26} & \textbf{31.91} & \textbf{13.08} & \textbf{62.44} & \textbf{86.18} & \textbf{58.13} & \textbf{35.40} \\ \rowcolor{LightCyan}
                                  & \textit{Improvement} & \textit{+6.91}  & \textit{+5.13}  & \textit{+7.35}  & \textit{+5.43}  & \textit{+5.17}  & \textit{+1.79}  & \textit{+5.16}  & \textit{+7.18}  \\ \hline
        \multirow{3}{*}{LEVEL\_2} & PointPillars          & 31.67          & 59.26          & 22.09          & 7.08           & 52.96          & 82.30          & 50.74          & 24.6           \\
                                  & SP + PointPillar      & \textbf{38.15} & \textbf{64.57} & \textbf{28.66} & \textbf{11.96} & \textbf{58.54} & \textbf{85.75} & \textbf{56.02} & \textbf{31.02} \\ \rowcolor{LightCyan}
                                  & \textit{Improvement} & \textit{+6.48}  & \textit{+5.31}  & \textit{+6.57}  & \textit{+4.88}  & \textit{+5.58}  & \textit{+3.45}  & \textit{+5.28}  & \textit{+6.42}  \\ \hline
        \multirow{3}{*}{LEVEL\_1} & PV-RCNN              & 55.16          & 76.68          & 47.96          & 27.59          & 74.01          & 91.39          & 70.94          & 49.51          \\
                                  & SPG+PV-RCNN          & \textbf{58.31} & \textbf{77.81}  & \textbf{51.65}  & \textbf{31.29} & \textbf{75.27} & \textbf{92.36} & \textbf{73.47} & \textbf{51.03} \\ \rowcolor{LightCyan}
                                  & \textit{Improvement} & \textit{+3.15}  & \textit{+1.13}  & \textit{+3.69}  & \textit{+3.70}  & \textit{+1.26}  & \textit{+0.97}  & \textit{+2.53}  & \textit{+1.52}   \\ \hline
        \multirow{3}{*}{LEVEL\_2} & PV-RCNN              & 45.81          & 71.31          & 38.83          & 20.52          & 64.69          & 88.95          & 64.80          & 37.37          \\
                                  & SPG + PV-RCNN          & \textbf{48.70} & \textbf{72.41} & \textbf{42.16} & \textbf{23.52} & \textbf{65.98} & \textbf{91.62} & \textbf{65.61} & \textbf{39.87} \\ \rowcolor{LightCyan}
                                  & \textit{Improvement} & \textit{+2.89}  & \textit{+1.10}  & \textit{+3.33}  & \textit{+3.00}  & \textit{+1.29}  & \textit{+2.67}  & \textit{+0.81}  & \textit{+2.50}   \\ \hline
        \end{tabular}
		\caption{The unsupervised domain adaptation vehicle detection results on both Waymo Open Dataset (OD) and Kirkland Dataset (Kirk). We show the vehicle 3D AP results in this table. The AP distance breakdowns are provided by the official evaluation tool.}
		\label{tb:uda_breakdown_v3d}
		\end{adjustwidth}
	\end{table*}
	
	\begin{table*}[!hbt]
      \begin{adjustwidth}{0pt}{0pt}
      \setlength\tabcolsep{7pt}
      \centering
       \begin{tabular}{c|c|cccc|cccc}
       \multicolumn{1}{l}{}                           & \multicolumn{1}{l}{}                                       & \multicolumn{4}{c}{Target Domain - Kirk}                                                                                                                                        & \multicolumn{4}{c}{Source Domain - OD}                                                                                                                     \\ 
        \hline
                                  &                      & \multicolumn{4}{c|}{Vehicle BEV AP (IoU = 0.7)}            & \multicolumn{4}{c}{Vehicle BEV AP (IoU = 0.7)}              \\
        Difficulty                & Method               & Overall        & 0-30m          & 30-50m         & 50-Inf         & Overall        & 0-30m          & 30-50m         & 50-Inf         \\ \hline
        \multirow{3}{*}{LEVEL\_1} & PointPillars          & 51.88          & 75.56          & 46.04          & 25.55          & 72.26          & 92.23          & 71.35          & 51.11          \\
                                  & SPG + PointPillars      & \textbf{60.44} & \textbf{80.89} & \textbf{53.73} & \textbf{38.24} & \textbf{77.63} & \textbf{93.39} & \textbf{75.96} & \textbf{61.16} \\ \rowcolor{LightCyan}
                                  & \textit{Improvement} & \textit{+8.56}  & \textit{+5.33}  & \textit{+7.69}  & \textit{+12.69} & \textit{+5.37}  & \textit{+1.16}  & \textit{+4.61}  & \textit{+10.05} \\ \hline
        \multirow{3}{*}{LEVEL\_2} & PointPillars          & 47.93          & 71.18          & 42.41          & 23.47          & 69.09          & 91.83          & 68.87          & 45.53          \\
                                  & SPG + PointPillars      & \textbf{56.94} & \textbf{77.13} & \textbf{49.99} & \textbf{35.04} & \textbf{74.90} & \textbf{93.06} & \textbf{73.96} & \textbf{54.51} \\ \rowcolor{LightCyan}
                                  & \textit{Improvement} & \textit{+9.01}  & \textit{+5.95}  & \textit{+7.58}  & \textit{+11.57} & \textit{+5.81}  & \textit{+1.23}  & \textit{+5.09}  & \textit{+8.98}  \\ \hline
        \multirow{3}{*}{LEVEL\_1} & PV-RCNN              & 70.38          & 84.27          & 65.31          & 52.98          & 85.13          & 95.99           & 84.02          & 72.19          \\
                                  & SPG + PV-RCNN          & \textbf{72.56} & \textbf{84.43} & \textbf{68.79} & \textbf{58.49} & \textbf{87.38} & \textbf{97.54} & \textbf{86.63} & \textbf{74.59} \\ \rowcolor{LightCyan}
                                  & \textit{Improvement} & \textit{+2.18}  & \textit{+0.16}  & \textit{+3.48}  & \textit{+5.51}  & \textit{+2.25}  & \textit{+1.55} & \textit{+2.61}  & \textit{+2.40}  \\ \hline
        \multirow{3}{*}{LEVEL\_2} & PV-RCNN              & 60.13          & 78.10          & 54.36          & 40.67          & 76.84          & 93.29          & 76.64          & 58.29          \\
                                  & SPG + PV-RCNN          & \textbf{62.03} & \textbf{78.86} & \textbf{56.47} & \textbf{44.94} & \textbf{78.05} & \textbf{94.45} & \textbf{80.25} & \textbf{59.56} \\ \rowcolor{LightCyan}
                                  & \textit{Improvement} & \textit{+1.90}  & \textit{+0.76}   & \textit{+2.11}  & \textit{+4.27}  & \textit{+1.21}  & \textit{+1.16} & \textit{+3.61}  & \textit{+1.27}  \\ \hline
        \end{tabular}
		\caption{The unsupervised domain adaptation vehicle detection results on both Waymo Open Dataset (OD) and Kirkland Dataset (Kirk). We show the vehicle BEV AP results in this table. The AP distance breakdowns are provided by the official evaluation tool.}
		\label{tb:uda_breakdown_vbev}
		\end{adjustwidth}
	\end{table*}
	
	\begin{table*}[!hbt]
      \begin{adjustwidth}{0pt}{0pt}
      \setlength\tabcolsep{7pt}
      \centering
       \begin{tabular}{c|c|cccc|cccc}
       \multicolumn{1}{l}{}                           & \multicolumn{1}{l}{}                                       & \multicolumn{4}{c}{Target Domain - Kirk}                                                                                                                                        & \multicolumn{4}{c}{Source Domain - OD}                                                                                                                     \\ 
        \hline
                                  &                      & \multicolumn{4}{c|}{Pedestrian 3D AP (IoU = 0.5)}          & \multicolumn{4}{c}{Pedestrian 3D AP (IoU = 0.5)}            \\
        Difficulty                & Method               & Overall        & 0-30m          & 30-50m         & 50-Inf         & Overall        & 0-30m          & 30-50m         & 50-Inf         \\ \hline
        \multirow{3}{*}{LEVEL\_1} & PointPillars          & 20.65          & 43.98          & 9.27           & 3.24           & 55.20          & 69.24          & 52.04          & 32.72          \\
                                  & SPG + PointPillars      & \textbf{23.72} & \textbf{50.19} & \textbf{9.11}  & \textbf{5.57}  & \textbf{56.06} & \textbf{69.32} & \textbf{53.12} & \textbf{34.73} \\ \rowcolor{LightCyan}
                                  & \textit{Improvement} & \textit{+3.07}  & \textit{+6.21}  & \textit{-0.16} & \textit{+2.33}  & \textit{+0.86}  & \textit{+0.08}  & \textit{+1.08}  & \textit{+2.01}  \\ \hline
        \multirow{3}{*}{LEVEL\_2} & PointPillars          & 17.66          & 40.67          & 7.40           & 2.32           & 51.33          & 65.85          & 49.32          & 29.29          \\
                                  & SPG + PointPillars      & \textbf{19.57} & \textbf{46.42} & \textbf{7.44}  & \textbf{3.99}  & \textbf{52.33} & \textbf{65.63} & \textbf{50.10} & \textbf{31.25} \\ \rowcolor{LightCyan}
                                  & \textit{Improvement} & \textit{+1.91}  & \textit{+5.75}  & \textit{+0.04}  & \textit{+1.67}  & \textit{+1.00}  & \textit{-0.22} & \textit{+0.78}  & \textit{+1.96}  \\ \hline
        \multirow{3}{*}{LEVEL\_1} & PV-RCNN              & 24.47          & 39.69          & 14.24          & 8.05           & 65.34          & 72.23          & 64.89          & 50.04          \\
                                  & SPG + PV-RCNN          & \textbf{30.82} & \textbf{48.04} & \textbf{18.80} & \textbf{13.39} & \textbf{66.93} & \textbf{73.55} & \textbf{66.60} & \textbf{50.82} \\ \rowcolor{LightCyan}
                                  & \textit{Improvement} & \textit{+6.35}  & \textit{+8.35}  & \textit{+4.56}  & \textit{+5.34}  & \textit{+1.59}  & \textit{+1.32}  & \textit{+1.71}  & \textit{+0.78}  \\ \hline
        \multirow{3}{*}{LEVEL\_2} & PV-RCNN              & 17.16          & 36.39          & 9.64           & 3.51           & 56.03          & 66.88          & 56.58          & 35.76          \\
                                  & SPG + PV-RCNN          & \textbf{22.05} & \textbf{44.07} & \textbf{12.91} & \textbf{5.77}  & \textbf{57.68} & \textbf{68.28} & \textbf{58.29} & \textbf{37.64} \\ \rowcolor{LightCyan}
                                  & \textit{Improvement} & \textit{+4.89}  & \textit{+7.68}  & \textit{+3.27}  & \textit{+2.26}  & \textit{+1.65}  & \textit{+1.40}  & \textit{+1.71}  & \textit{+1.88}  \\ \hline
        \end{tabular}
		\caption{The unsupervised domain adaptation pedestrian detection results on both Waymo Open Dataset (OD) and Kirkland Dataset (Kirk). We show the pedestrian 3D AP results in this table. The AP distance breakdowns are provided by the official evaluation tool.}
		\label{tb:uda_breakdown_p3d}
		\end{adjustwidth}
	\end{table*}
	
	\begin{table*}[!hbt]
      \begin{adjustwidth}{0pt}{0pt}
      \setlength\tabcolsep{7pt}
      \centering
       \begin{tabular}{c|c|cccc|cccc}
       \multicolumn{1}{l}{}                           & \multicolumn{1}{l}{}                                       & \multicolumn{4}{c}{Target Domain - Kirk}                                                                                                                                        & \multicolumn{4}{c}{Source Domain - OD}                                                                                                                     \\ 
        \hline
                                  &                      & \multicolumn{4}{c|}{Pedestrian BEV AP (IoU = 0.5)}         & \multicolumn{4}{c}{Pedestrian BEV AP (IoU = 0.5)}           \\
        Difficulty                & Method               & Overall        & 0-30m          & 30-50m         & 50-Inf         & Overall        & 0-30m          & 30-50m         & 50-Inf         \\ \hline
        \multirow{3}{*}{LEVEL\_1} & PointPillars         & 22.33          & 45.00          & 10.50          & 3.49           & 63.82          & 76.33          & 61.90          & 42.81          \\
                                  & SPG + PointPillars   & \textbf{24.83} & \textbf{51.44} & \textbf{10.80} & \textbf{5.71}  & \textbf{64.66} & \textbf{76.11} & \textbf{62.69} & \textbf{44.98} \\ \rowcolor{LightCyan}
                                  & \textit{Improvement} & \textit{+2.50}  & \textit{+6.44}  & \textit{+0.30}  & \textit{+2.22}  & \textit{+0.84}  & \textit{-0.22} & \textit{+0.79}  & \textit{+2.17}  \\ \hline
        \multirow{3}{*}{LEVEL\_2} & PointPillars         & 18.40          & 41.63          & 8.58           & 2.49           & 60.13          & 73.34          & 58.77          & 38.83          \\
                                  & SPG + PointPillars   & \textbf{20.67} & \textbf{47.56} & \textbf{8.98}  & \textbf{4.11}  & \textbf{60.93} & \textbf{72.94} & \textbf{59.54} & \textbf{41.11} \\ \rowcolor{LightCyan}
                                  & \textit{Improvement} & \textit{+2.27}  & \textit{+5.93}  & \textit{+0.40}  & \textit{+1.62}  & \textit{+0.80}  & \textit{-0.40} & \textit{+0.77}  & \textit{+2.28}  \\ \hline
        \multirow{3}{*}{LEVEL\_1} & PV-RCNN              & 25.39          & 40.23          & 14.72          & 9.76           & 70.35          & 76.22          & 70.49          & 56.77          \\
                                  & SPG + PV-RCNN        & \textbf{31.92} & \textbf{49.06} & \textbf{19.87} & \textbf{14.87} & \textbf{70.37} & \textbf{75.86} & \textbf{72.29} & \textbf{57.47} \\ \rowcolor{LightCyan}
                                  & \textit{Improvement} & \textit{+6.53}  & \textit{+8.83}  & \textit{+5.15}  & \textit{+5.11}  & \textit{+0.02}  & \textit{-0.36} & \textit{+1.80}  & \textit{+0.70}  \\ \hline
        \multirow{3}{*}{LEVEL\_2} & PV-RCNN              & 17.88          & 36.89          & 9.97           & 4.23           & 60.81          & 69.22          & 61.86          & 41.32          \\
                                  & SPG + PV-RCNN        & \textbf{22.65} & \textbf{44.57} & \textbf{13.48} & \textbf{6.38}  & \textbf{60.88} & \textbf{70.62} & \textbf{63.65} & \textbf{43.27} \\ \rowcolor{LightCyan}
                                  & \textit{Improvement} & \textit{+4.77}  & \textit{+7.68}  & \textit{+3.51}  & \textit{+2.15}  & \textit{+0.07}  & \textit{+1.40}  & \textit{+1.79}  & \textit{+1.95}  \\ \hline
        \end{tabular}
		\caption{The unsupervised domain adaptation pedestrian detection results on both Waymo Open Dataset (OD) and Kirkland Dataset (Kirk). We show the pedestrian BEV AP results in this table. The AP distance breakdowns are provided by the official evaluation tool.}
		\label{tb:uda_breakdown_pbev}
		\end{adjustwidth}
	\end{table*}
	
\section{More Results on KITTI}
    \label{sec:More Results on KITTI}
    We provide more 3D object detection results on KITTI. There are two commonly used metric standards for evaluating the detection performance: 1) R11, where the AP is evaluated with 11 recall positions; 2) R40, where the AP is evaluated with 40 recall positions. In addition to the improvement on car and pedestrian detection, SPG also significantly boosts the performance in cyclist detection. Based on R11, Table \ref{tb:kitti_r113d} and Table \ref{tb:kitti_r11bev} show the results in 3D AP and BEV AP for three object types, respectively. 
    Based on R40, Table \ref{tb:kitti_r403d} and Table \ref{tb:kitti_r40bev} show the results in 3D AP and BEV AP for three object types, respectively.
    
    We show more comparisons on the KITTI test set in Table \ref{tb:kitti_test}. 
    
    \begin{table*}[!hbt]
       \begin{adjustwidth}{0pt}{0pt}
      \setlength\tabcolsep{10pt}
      \centering
       \begin{tabular}{c|ccc|ccc|ccc}
        \hline
                             & \multicolumn{3}{c|}{Car - 3D AP (R11)}                   & \multicolumn{3}{c|}{Pedestrian - 3D AP (R11)}            & \multicolumn{3}{c}{Cyclist - 3D AP (R11)}                \\
        Method               & Easy           & Mod.           & Hard           & Easy           & Mod.           & Hard           & Easy           & Mod.           & Hard           \\ \hline
        PointPillars\cite{lang2019pointpillars}          & 86.46          & 77.28          & 74.65          & 57.75          & 52.29          & 47.90          & 80.05          & 62.68          & 59.70          \\
        SPG + PointPillars      & \textbf{87.98} & \textbf{78.54} & \textbf{77.32} & \textbf{59.91} & \textbf{54.58} & \textbf{50.34} & \textbf{81.58} & \textbf{65.70} & \textbf{62.28} \\ \rowcolor{LightCyan}
        \textit{Improvement} & \textit{+1.52}  & \textit{+1.26}  & \textit{+2.67}  & \textit{+2.16}  & \textit{+2.29}  & \textit{+2.44}  & \textit{+1.53}  & \textit{+3.02}  & \textit{+2.58}  \\ \hline
        PVRCNN\cite{shi2020pv}               & 89.35          & 83.69          & 78.70          & 64.60          & 57.90          & 53.23          & 85.22          & 70.47          & 65.75          \\
        SPG + PVRCNN           & \textbf{89.81} & \textbf{84.45} & \textbf{79.14} & \textbf{69.04} & \textbf{62.18} & \textbf{56.77} & \textbf{86.82} & \textbf{73.35} & \textbf{69.30} \\ \rowcolor{LightCyan}
        \textit{Improvement} & \textit{+0.46}  & \textit{+0.76}  & \textit{+0.44}  & \textit{+4.44}  & \textit{+4.28}  & \textit{+3.54}  & \textit{+1.60}  & \textit{+2.88}  & \textit{+3.55}  \\ \hline
        \end{tabular}
		\caption{Result comparisons on the KITTI validation set. The results are evaluated by the Average Precision with 11 recall positions. The baseline detectors, PointPillars and PV-RCNN, are directly evaluated by using the checkpoints released by \cite{shi2020pv,openpcdet2020}. 
		}
		\label{tb:kitti_r113d}
		\end{adjustwidth}
	\end{table*}
	
	\begin{table*}[!hbt]
       \begin{adjustwidth}{0pt}{0pt}
      \setlength\tabcolsep{10pt}
      \centering
       \begin{tabular}{c|ccc|ccc|ccc}
        \hline
                             & \multicolumn{3}{c|}{Car - BEV AP (R11)}                  & \multicolumn{3}{c|}{Pedestrian - BEV AP (R11)}           & \multicolumn{3}{c}{Cyclist - BEV AP (R11)}               \\
        Method               & Easy           & Mod.           & Hard           & Easy           & Mod.           & hard           & Easy           & Mod.           & Hard           \\ \hline
        PointPillars\cite{lang2019pointpillars}           & 89.65          & 87.17          & 84.37          & 61.63          & 56.27          & 52.60          & 82.27          & 66.25          & 62.64          \\
        SPG + PointPillars    & \textbf{90.07} & \textbf{88.00} & \textbf{86.63} & \textbf{65.16} & \textbf{59.86} & \textbf{56.07} & \textbf{86.02} & \textbf{71.93} & \textbf{65.69} \\ \rowcolor{LightCyan}
        \textit{Improvement} & \textit{+0.42}  & \textit{+0.83}  & \textit{+2.26}  & \textit{+3.53}  & \textit{+3.59}  & \textit{+3.47}  & \textit{+3.75}  & \textit{+5.68}  & \textit{+3.05}  \\ \hline
        PVRCNN\cite{shi2020pv}               & 90.09          & 87.90          & 87.41          & 67.01          & 61.38          & 56.10          & 86.79          & 73.55          & 69.69          \\
        SPG + PVRCNN         & \textbf{90.41} & \textbf{88.49} & \textbf{87.74} & \textbf{71.19} & \textbf{64.37} & \textbf{59.88} & \textbf{92.54} & \textbf{74.43} & \textbf{70.99} \\ \rowcolor{LightCyan}
        \textit{Improvement} & \textit{+0.32}  & \textit{+0.59}  & \textit{+0.33}  & \textit{+4.18}  & \textit{+2.99}  & \textit{+3.78}  & \textit{+5.75}  & \textit{+0.88}  & \textit{+1.30}  \\ \hline
        \end{tabular}
		\caption{Result comparisons on the KITTI validation set. The results are evaluated by the Average Precision with 11 recall positions. The baseline detectors, PointPillars and PV-RCNN, are directly evaluated by using the checkpoints released by \cite{shi2020pv,openpcdet2020}. 
		}
		\label{tb:kitti_r11bev}
		\end{adjustwidth}
	\end{table*}
	
	\begin{table*}[!hbt]
       \begin{adjustwidth}{0pt}{0pt}
      \setlength\tabcolsep{10pt}
      \centering
       \begin{tabular}{c|ccc|ccc|ccc}
        \hline
                             & \multicolumn{3}{c|}{Car - 3D AP (R40)}             & \multicolumn{3}{c|}{Pedestrian - 3D AP  (R40)}     & \multicolumn{3}{c}{Cyclist - 3d AP  (R40)}         \\
        Method               & Easy           & Mod.           & Hard           & Easy           & Mod.           & Hard           & Easy           & Mod.           & Hard           \\ \hline
        PointPillars\cite{lang2019pointpillars}           & 87.75          & 78.39          & 75.18          & 57.30          & 51.41          & 46.87          & 81.57          & 62.94          & 58.98          \\
        SPG+PointPillars      & \textbf{89.77} & \textbf{81.36} & \textbf{78.85} & \textbf{59.65} & \textbf{53.55} & \textbf{49.24} & \textbf{83.27} & \textbf{66.11} & \textbf{61.99} \\ \rowcolor{LightCyan}
        \textit{Improvement} & \textit{+2.02}  & \textit{+2.97}  & \textit{+3.67}  & \textit{+2.35}  & \textit{+2.14}  & \textit{+2.37}  & \textit{+1.70}  & \textit{+3.17}  & \textit{+3.01}  \\ \hline
        PVRCNN\cite{shi2020pv}               & 92.10          & 84.36          & 82.48          & 64.26          & 56.67          & 51.91          & 88.88          & 71.95          & 66.78          \\
        SPG+PVRCNN           & \textbf{92.53} & \textbf{85.31} & \textbf{82.82} & \textbf{69.66} & \textbf{61.80} & \textbf{56.39} & \textbf{91.75} & \textbf{74.35} & \textbf{69.49} \\ \rowcolor{LightCyan}
        \textit{Improvement} & \textit{+0.43}  & \textit{+0.95}  & \textit{+0.34}  & \textit{+5.40}  & \textit{+5.13}  & \textit{+4.48}  & \textit{+2.87}  & \textit{+2.40}  & \textit{+2.71}  \\ \hline
        \end{tabular}
		\caption{Result comparisons on the KITTI validation set. The results are evaluated by the Average Precision with 40 recall positions. The baseline detectors, PointPillars and PV-RCNN, are directly evaluated by using the checkpoints released by \cite{shi2020pv,openpcdet2020}.
		}
		\label{tb:kitti_r403d}
		\end{adjustwidth}
	\end{table*}
	
	\begin{table*}[!hbt]
       \begin{adjustwidth}{0pt}{0pt}
      \setlength\tabcolsep{10pt}
      \centering
       \begin{tabular}{c|ccc|ccc|ccc}
        \hline
         & \multicolumn{3}{c|}{Car - BEV AP (R40)}            & \multicolumn{3}{c|}{Pedestrian - BEV AP  (R40)}    & \multicolumn{3}{c}{Cyclist - BEV AP  (R40)}        \\
        Method               & Easy           & Mod.           & Hard           & Easy           & Mod.           & Hard           & Easy           & Mod.           & Hard           \\ \hline
        PointPillars\cite{lang2019pointpillars}           & 92.03          & 88.05          & 86.66          & 61.59          & 56.01          & 52.04          & 85.27          & 66.34          & 62.36          \\
        SPG + PointPillars    & \textbf{94.38} & \textbf{89.92} & \textbf{87.97} & \textbf{65.38} & \textbf{59.48} & \textbf{55.32} & \textbf{90.29} & \textbf{71.43} & \textbf{66.96} \\ \rowcolor{LightCyan}
        \textit{Improvement} & \textit{+2.35}  & \textit{+1.87}  & \textit{+1.31}  & \textit{+3.79}  & \textit{+3.47}  & \textit{+3.28}  & \textit{+5.02}  & \textit{+5.09}  & \textit{+4.60}  \\ \hline
        PVRCNN\cite{shi2020pv}               & 93.02          & 90.33          & 88.53          & 67.97          & 60.52          & 55.80          & 91.02          & 74.54          & 69.92          \\
        SPG + PVRCNN         & \textbf{94.99} & \textbf{91.11} & \textbf{88.86} & \textbf{71.79} & \textbf{64.50} & \textbf{59.51} & \textbf{93.62} & \textbf{76.45} & \textbf{71.64} \\ \rowcolor{LightCyan}
        \textit{Improvement} & \textit{+1.97}  & \textit{+0.78}  & \textit{+0.33}  & \textit{+3.82}  & \textit{+3.98}  & \textit{+3.71}  & \textit{+2.60}  & \textit{+1.91}  & \textit{+1.72}  \\ \hline
        \end{tabular}
		\caption{Result comparisons on the KITTI validation set. The results are evaluated by the Average Precision with 40 recall positions. The baseline detectors, PointPillars and PV-RCNN, are directly evaluated by using the checkpoints released by \cite{shi2020pv,openpcdet2020}.
		}
		\label{tb:kitti_r40bev}
		\end{adjustwidth}
	\end{table*}
	
	 \begin{table*}[!hbt]
	    \begin{adjustwidth}{0pt}{0pt}
            \setlength\tabcolsep{10pt}
            \centering
            \begin{tabular}{c|cc|cccc}
            \hline
            &                                     &              & \multicolumn{4}{c}{Car - 3D AP}                                  \\
            Method                                                  & Reference                           & Modality     & Easy           & Mod.           & Hard           & Avg.           \\ \hline
            F-PointNet\cite{qi2018frustum}        & \small CVPR 2018     & LIDAR \& RGB & 82.19          & 69.79          & 60.59          & 70.86          \\
            AVOD-FPN\cite{ku2018joint}        & \small IROS 2018     & LIDAR \& RGB & 83.07         & 71.76          & 65.73          & 73.52          \\
            F-ConvNet\cite{wang2019frustum}        & \small IROS 2019     & LIDAR \& RGB & 87.36          & 76.39          & 66.69          & 76.81          \\
            UberATG-MMF\cite{liang2019multi}       & \small CVPR 2019     & LIDAR \& RGB & 88.40          & 77.43          & 70.22          & 78.68          \\
            EPNet\cite{huang2020epnet}             & \small ECCV 2020     & LiDAR \& RGB & 89.81          & 79.28          & 74.59          & 81.23          \\
            CLOCs\_PVCas\cite{pang2020clocs}       & \small IROS 2020     & LiDAR \& RGB & 88.94          & 80.67          & 77.15          & 82.25          \\
            3D-CVF\cite{yoo20203d}                 & \small ECCV 2020     & LiDAR \& RGB & 89.20          & 80.05          & 73.11          & 80.79          \\ \hline
            SECOND\cite{yan2018second}    & \small Sensors 2018     & LiDAR        & 83.34          & 72.55          & 65.82          & 73.90          \\
            PointPillars\cite{lang2019pointpillars}    & \small CVPR 2019    & LiDAR        & 82.58          & 74.31          & 68.99          & 75.30          \\
            PointRCNN\cite{shi2019pointrcnn}    & \small CVPR 2019    & LiDAR        & 86.96          & 76.50          & 71.39          & 77.77          \\
            3D IoU Loss\cite{zhou2019iou}    & \small 3DV 2019    & LiDAR        & 86.16          & 75.64          & 70.70          & 78.28          \\
            Fast Point R-CNNs\cite{chen2019fast}    & \small ICCV 2019    & LiDAR        & 85.29          & 77.40          & 70.24          & 77.64          \\
            STD\cite{yang2019std}               & \small ICCV 2019    & LiDAR        & 87.95        & 79.71             & 75.09          & 80.91         \\
            SegVoxelNet\cite{yi2020segvoxelnet}    & \small ICRA 2020     & LiDAR        & 86.04          & 76.13          & 70.76          & 77.64          \\
            SARPNET\cite{ye2020sarpnet}            & \small Neuro Computing 2019 & LiDAR        & 85.63          & 76.64          & 71.31          & 77.86          \\
            HRI-VoxelFPN\cite{yi2020segvoxelnet}   & \small Sensor 2020   & LiDAR        & 85.63          & 76.70          & 69.44          & 77.26          \\
            HotSpotNet\cite{chen2020object}        & \small ECCV 2020     & LiDAR        & 87.60          & 78.31          & 73.34          & 79.75          \\
            PartA$^2$\cite{9018080}                   & \small TPAMI 2020    & LiDAR        & 87.81          & 78.49          & 73.51          & 79.94          \\
            SERCNN\cite{Zhou_2020_CVPR}          & \small CVPR 2020     & LiDAR        & 87,74          & 78.96          & 74.14          & 51.03          \\
            Point-GNN\cite{shi2020point}           & \small CVPR 2020     & LiDAR        & 88.33          & 79.47          & 72.29          & 80.03          \\
            3DSSD\cite{yang20203dssd}              & \small CVPR 2020     & LiDAR        & 88.36          & 79.57          & 74.55          & 80.83         \\
            SA-SSD\cite{he2020sassd}               & \small CVPR 2020     & LiDAR        & 88.75          & 79.79          & 74.16          & 80.90          \\
            CIA-SSD\cite{zheng2020ciassd}          & \small AAAI 2021     & LiDAR        & 89.59          & 80.28          & 72.87          & 80.91          \\
            Asso-3Ddet\cite{du2020associate}       & \small CVPR 2020     & LiDAR        & 85.99          & 77.40          & 70.53          & 77.97          \\
            Voxel R-CNN\cite{deng2020voxel}        & \small AAAI 2021     & LiDAR        & \textbf{90.90} & 81.62          & 77.06          & 83.19          \\ \hline
            PV-RCNN\cite{shi2020pv}                & \small CVPR 2020     & LiDAR        & 90.25          & 81.43          & 76.82          & 82.83          \\
            SPG+PV-RCNN (Ours)                                             & -                                   & LiDAR        & 90.49          & \textbf{82.13} & \textbf{78.88} & \textbf{83.83} \\ \hline
            \end{tabular}
    		\caption{Car detection result comparisons on the KITTI test set. The results are evaluated by the Average Precision with 40 recall positions on the KITTI benchmark website. We compare with the leader board front runner detectors that are associated with conferences or journals released before our submission. The Avg. AP is calculated by averaging over the APs of Easy, Mod. and Hard. difficulty levels.
    		}
    		\label{tb:kitti_test}
    		\end{adjustwidth}
    \end{table*}

\section{More Visualization of Semantic Point Generation}
    \label{sec:More Visualization of Semantic Point Generation}
\begin{figure*}[!hbt]
        \vspace{-10pt}
        \centering
        \begin{subfigure}{\textwidth}
            \includegraphics[width=1.0\linewidth]{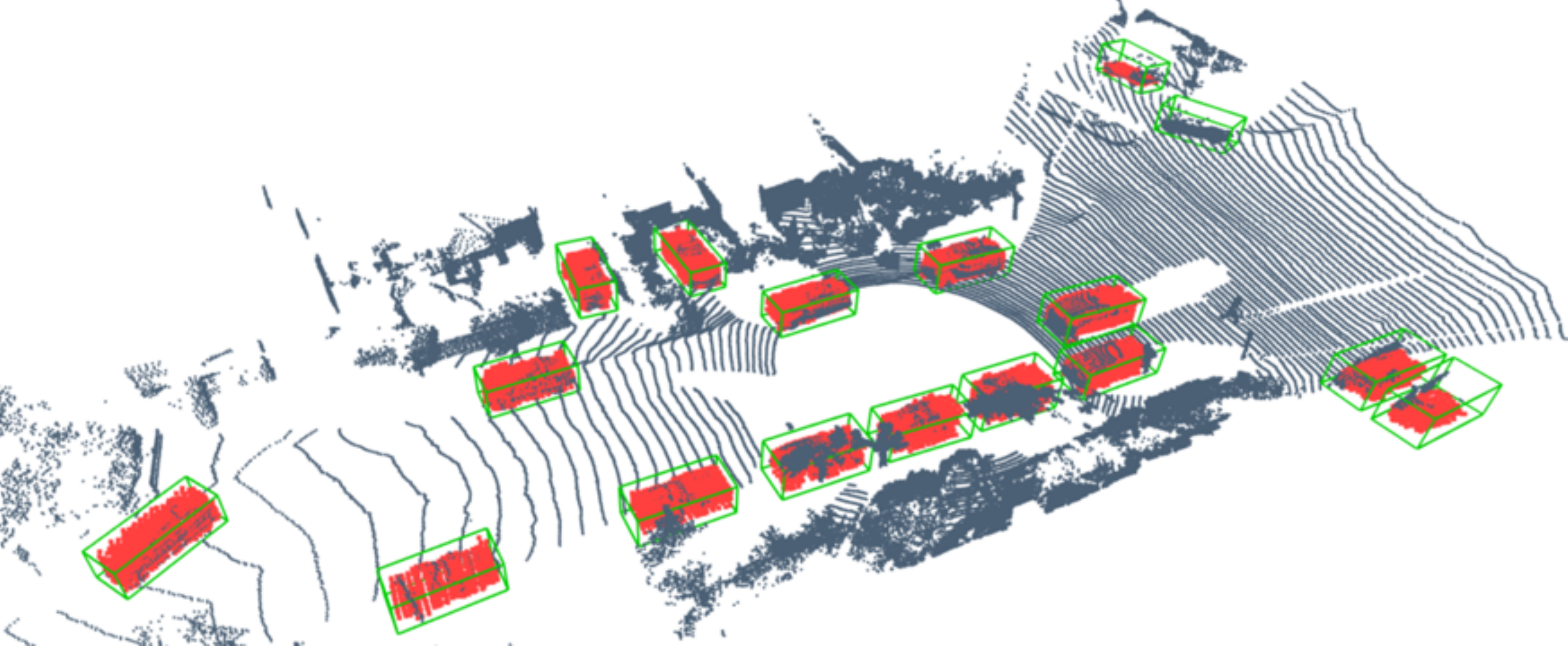} 
            \vspace*{10pt}
        \end{subfigure}
        \begin{subfigure}{\textwidth}
            \includegraphics[width=1.0\linewidth]{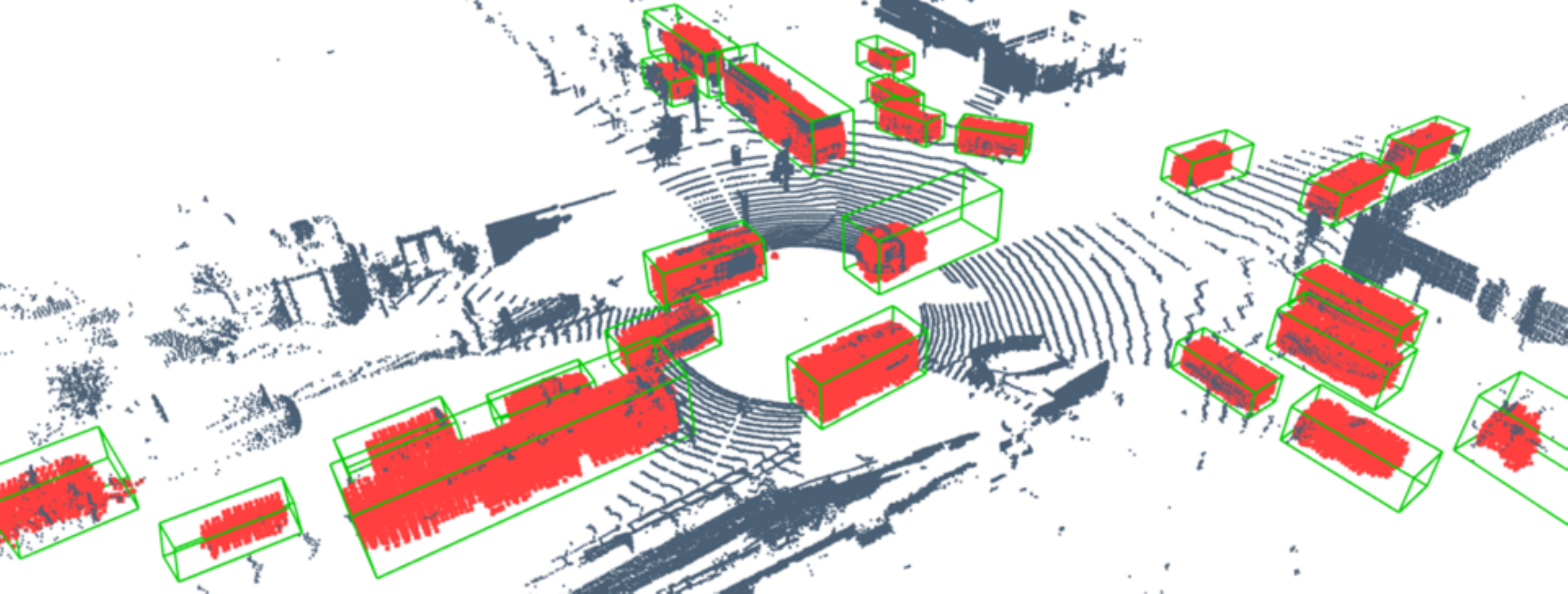} 
            \vspace*{10pt}
        \end{subfigure}
        \begin{subfigure}{\textwidth}
            \includegraphics[width=1.0\linewidth]{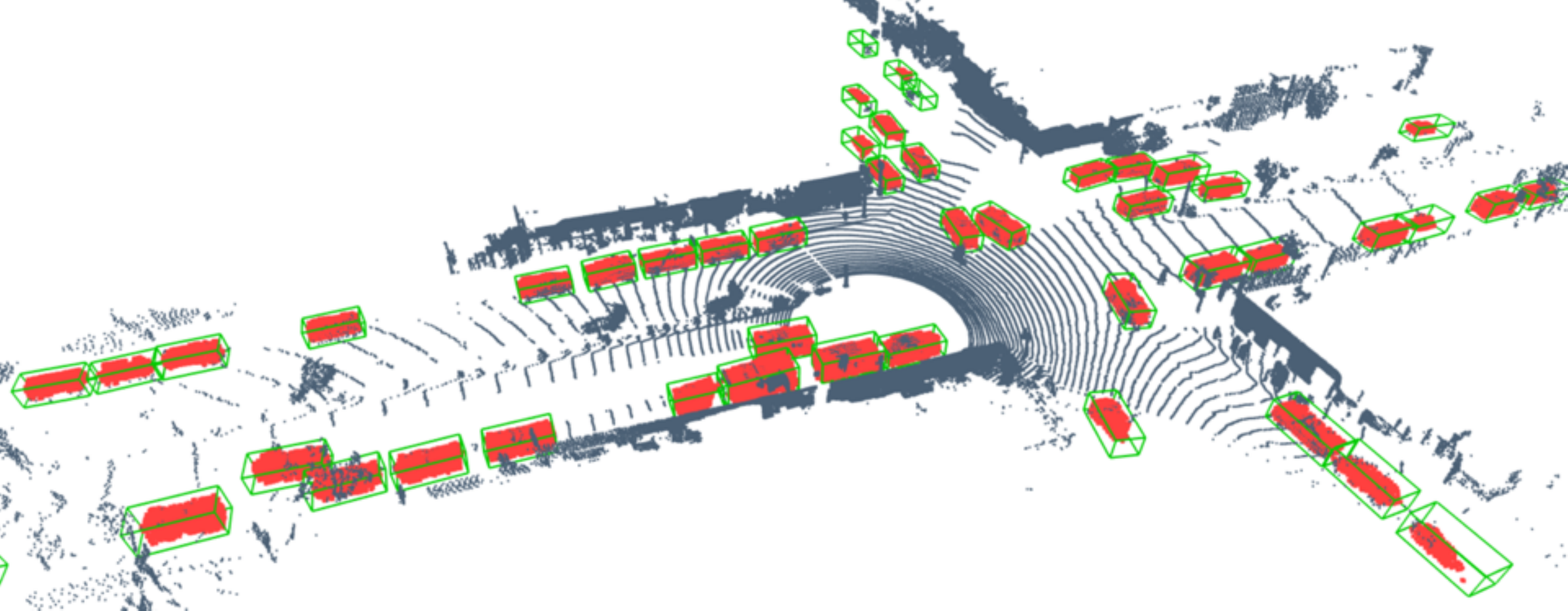} 
        \end{subfigure}
        \caption{More visualization of generated semantic points. The grey points are original raw points. The red points are the generated semantic points. The green boxes are the predicted bounding boxes.}
        \label{fig:morevis}
    \end{figure*}

    In Figure \ref{fig:morevis}, we illustrate more augmented point clouds, where the raw points are rendered in the grey color and the generated semantic points are highlighted in red.
    \begin{figure*}[!hbt]
        \centering
        \begin{subfigure}{\textwidth}
            \includegraphics[width=1.0\linewidth]{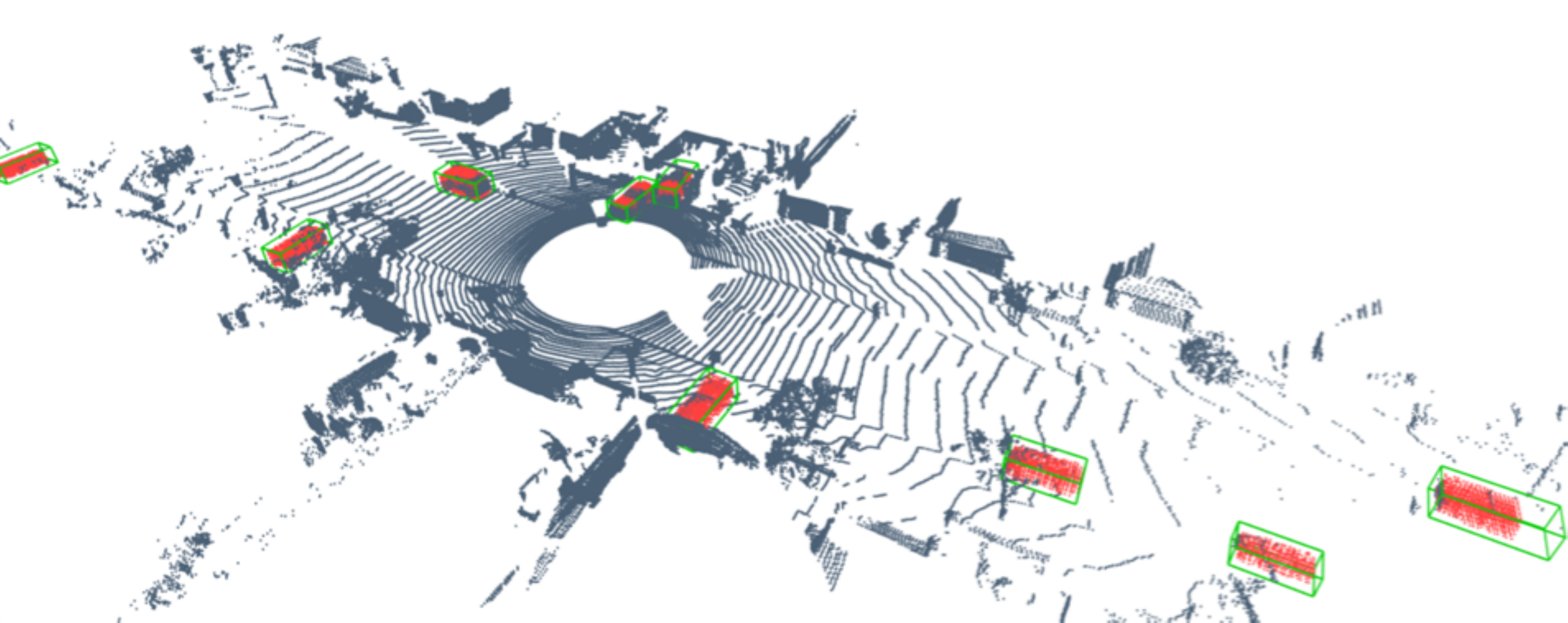} 
        \end{subfigure}
    \end{figure*}

\end{appendices}
\end{document}